\input ./style/arxiv-general.cfg
\documentclass[aos,MSNbibl,seceqn,secthm,dvips]{arximspdf}
\makeatletter
   \@ifpackageloaded{graphicx}{}{\usepackage{graphicx}}
\makeatother

\doi{10.1214/14-AOS1285}
\volume{43}
\issue{3}
\pubyear{2015}
\firstpage{962}
\lastpage{990}
\docsubty{FLA}

\makeatletter
\let\oldrho\rho
\renewcommand{\rho}{\oldrho_n}
\let\olda\alpha
\renewcommand{\alpha}{\olda_n}
\let\oldb\beta
\renewcommand{\beta}{\oldb_n}
\let\oldg\gamma
\renewcommand{\gamma}{\oldg_n}
\newtheorem{lem}[thm]{Lemma}
\newproclaim{defn}{Definition}[section]
\newtheorem{theorem}{Theorem}[section]

\newtheorem{thmc}{Theorem}[section]
\newproclaim{assumption}[thma]{Assumption}
\newtheorem{cor}[thmc]{Corollary}
\newproclaim{remark}{Remark}
\makeatother

\begin{document}
\begin{frontmatter}

\title{Role of normalization in spectral clustering for~stochastic blockmodels}
\runtitle{Role of normalization in spectral clustering}

\begin{aug}
\author[A]{\fnms{Purnamrita}~\snm{Sarkar}\corref{}\thanksref{T1}\ead[label=e1]{purna.sarkar@austin.utexas.edu}}
\and
\author[B]{\fnms{Peter J.}~\snm{Bickel}\ead[label=e2]{bickel@stat.berkeley.edu}}
\runauthor{P. Sarkar and P. J. Bickel}
\affiliation{University of Texas, Austin and University of California,
Berkeley}
\address[A]{Department of Statistics and Data Science\\
University of Texas, Austin\\
Austin, Texas 78712\\
USA\\
\printead{e1}}
\address[B]{Department of Statistics\\
University of California, Berkeley\\
Berkeley, California 94720\\
USA\\
\printead{e2}}
\end{aug}
\thankstext{T1}{Supported in part by NSF FRG Grant DMS-11-60319.}

%
\received{\smonth{10} \syear{2013}}
%
\revised{\smonth{11} \syear{2014}}

%
\begin{abstract}
Spectral clustering is a technique that clusters elements using the top
few eigenvectors of their (possibly normalized) similarity matrix.
The quality of spectral clustering is closely tied to the convergence
properties of these principal eigenvectors.
This rate of convergence has been shown to be identical for both the
normalized and unnormalized variants in recent random matrix theory literature.
However, normalization for spectral clustering is commonly believed to
be beneficial [\textit{Stat. Comput.} \textbf{17} (2007)
395--416]. Indeed, our experiments show that normalization improves
prediction accuracy.
In this paper, for the popular stochastic blockmodel, 
we theoretically show that normalization shrinks the spread of points
in a class by a constant fraction under a broad parameter regime.
As a byproduct of our work, we also obtain sharp deviation bounds of
empirical principal eigenvalues of graphs generated from a stochastic
blockmodel.
\end{abstract}

%
\end{frontmatter}

\section{Introduction}
\label{secintro}
Networks appear in many real-world problems. Any dataset of
co-occurrences or relationships between pairs of entities can be
represented as a network. For example, the Netflix data can be thought
of as a giant bipartite network between customers and movies, where
edges are formed via ratings. Facebook is a network of friends, where
edges represent who knows whom. Weblogs link to other blogs and give
rise to blog networks. Networks can also be implicit; for example, in
machine learning they are often built by computing pairwise
similarities between entities.

Many problems in machine learning and statistics are centered around
community detection. Viral marketing functions by understanding how
information propagates through friendship networks, and community
detection is key to this. Link farms in the World Wide Web are
basically malicious tightly connected clusters of webpages which
exploit web-search algorithms to increase their rank.
These need to be identified and removed so that search results are
authentic and do not mislead users.

Spectral clustering \cite{donath1973,Fiedler1973} is a widely
used network clustering algorithm. The main idea is to first represent
the $i$th entity by a $k$-dimensional vector obtained by concatenating
the $i$th elements of the top $k$ eigenvectors of a graph, and then
cluster this lower-dimensional representation. We will refer to this as
the spectral representation.
Due to its computational ease and competitive performance, emerging
application areas of spectral clustering range widely from parallel
computing \cite{Hendrickson1995}, CAD (computer aided design) \cite
{HagenK92}, parallel sparse matrix factorization \cite{Pothen1990} to
image segmentation \cite{Shimalik}, general clustering problems in
machine learning \cite{ng01onspectral} and most recently, to fitting
and classification using network blockmodels \cite
{rohechatterjiyu,priebejasa2012}.


A stochastic blockmodel is a widely used generative model for networks
with labeled nodes \cite{holland-leinhardt,rohechatterjiyu}. It
assigns nodes to $k$ different classes and
forces all nodes in the same class to be stochastically equivalent. For
example, in a two-class stochastic blockmodel, any pair of nodes
belonging to different classes link with probability $\gamma$ (a
deterministic quantity possibly dependent on the size of the graph,
i.e., $n$), whereas any pair belonging to class one (two) link with
probability $\alpha$ ($\beta$).

Recently the consistency properties of spectral clustering in the
context of stochastic blockmodels have attracted a significant amount
of attention.
Rohe, Chatterjee and Yu
\cite{rohechatterjiyu} showed that, under general
conditions, for a sequence of normalized graphs with growing size
generated from a stochastic blockmodels, spectral clustering yields the
correct clustering in the limit. In a subsequent paper,
Sussman et~al.
\cite{priebejasa2012} showed that an analogous statement holds for an
unnormalized sequence of graphs. For finite $k$, the above results can
also be obtained using direct applications of results from \cite{Oliveira2010}.

This prior theoretical work does not distinguish between normalized
and unnormalized spectral clustering, and hence cannot be used to
support the common practice of normalizing matrices for spectral clustering.
In this paper, we present both theoretical arguments and empirical
results to give a quantitative argument showing that normalization
improves the quality of clustering.
While existing work \cite{rohechatterjiyu,priebejasa2012} bounds
the classification accuracy, we do not take this route, since upper
bounds can not be used to compare two methods. Instead, we focus on the
variance within a class under the spectral representation using the top
$k$ eigenvectors. In this representation, by virtue of stochastic
equivalence, points are identically distributed around their respective
class centers. Hence the empirical variance can be computed using the
average squared distance of points from their class center.


In this setting, the distance between the class centers can be thought
of as bias; we show that this distance approaches the same
deterministic quantity with or without normalization. Surprisingly, we
also prove that normalization reduces the variance of points in a class
by a constant fraction for a large parameter regime. So normalization
does not change the bias, but shrinks the variance asymptotically.
However, our results also indicate that the variance of points in a
class increases as the graph gets sparser; hence methods which reduce
the within-class variance are desired.

A simple consequence of our result is that in the completely
associative case ($\gamma=0$) as well as the completely dissociative
case ($\alpha=\beta=0$), the variance of the spectral embedding within
a class is asymptotically four times less when the matrix is
normalized. While the completely associative case is on its own
uninteresting, we build the proof of the general case using similar
ideas and techniques.


Our results indicate that normalization has a clear edge when the
parameters are close to the completely associative or completely
dissociative settings. These seemingly easy to cluster regimes can be
relatively difficult in sparse networks. Of course, as $n$ grows, both
methods have enough data to distinguish between the clusters and behave
similarly. But for small and sparse graphs, it is indeed an important regime.

Sussman et~al.
\cite{priebejasa2012} present a parameter setting where
normalization is shown to hurt classification accuracy empirically. We
show that this is but a partial picture; and in fact there is a large
parameter regime where spectral clustering with normalized matrices
yields tighter and hence better clusters.

Using quantifiable link prediction experiments on real world graphs
and classifications tasks in labeled simulated graphs, we show that
normalization leads to better classification accuracies for the regime
dictated by our theory, and yields higher link prediction accuracy on
sparse real world graphs.

We conclude the introduction with a word of caution. Our asymptotic
theory is valid in the degree regime where networks are connected with
probability approaching one. However, finite sparse networks can have
disconnected or weakly connected small components, in the presence of
which, the normalized method returns uninformative principal
eigenvectors with support on the small components. This makes
classification worse compared to the unnormalized method, whose
principal eigenvectors are informative in spite of having high variance.
Hence, our asymptotic results should be used only as a guidance for
finite $n$, not as a hard rule. We deal with this problem by removing
low degree nodes and performing experiments on the giant component of
the remaining network.

\section{Preliminaries and import of previous work}
\label{secprob-def}
In this paper we will only work with two class blockmodels. 
Given a binary $n\times2$ class membership matrix~$Z$, the edges of
the network are simply outcomes of $n\choose2$ independent\vspace*{1pt} Bernoulli
coin flips. The stochastic blockmodel ensures stochastic equivalence of
nodes within the same block; that is, all nodes within the same block
have identical probability of linking with other nodes in the graph.

Thus the conditional expectation matrix $P$ of the adjacency matrix $A$
can be described by three probabilities, namely $\alpha,\beta,\gamma$;
where $\alpha$ and $\beta$ denote the probabilities of connecting
within the first and second classes ($C_1$ and $C_2$), respectively, and
$\gamma$ denotes the probability of connecting across two classes. All
statements in this paper are conditioned on $\alpha,\beta,\gamma$
and $Z$.

\begin{defn}[(A stochastic blockmodel)]
\label{defSBM1}
Let $Z\in\{0,1\}^{n\times2}$ be a fixed and unknown matrix of class
memberships such that every row has exactly one $1$, and the first and
second columns have $n\pi$ and $n(1-\pi)$ ones, respectively. A~stochastic blockmodel with parameters $(\alpha,\beta,\gamma,Z)$
generates symmetric graphs with adjacency matrix $A$ such that,
$P(A(i,i)=1)=0$, $\forall i$. For $i> j$, $A_{ij}=A_{ji}$ are
independent with $P(A_{ij}=1|Z)=P_{ij}$,
where $P$ is symmetric with $P_{ij}=\alpha$ for $i,j\in C_1$, $\gamma$
for $i\in C_1,j\in C_2$ and $\beta$ for $i,j\in C_2$.
\end{defn}

For ease of exposition we will assume that the rows and columns of $A$
are permuted such that all elements of the same class are grouped together.
We have $P:=E[A|Z]$. Clearly, $P$ is a blockwise constant matrix
with zero diagonal by construction.

We use a parametrization similar to that in \cite
{bickel2009nonparametric} to allow for decaying edge probabilities as
$n$ grows. Formally $\alpha$, $\beta$ and $\gamma$ are proportional to
a common rate variable $\rho$ where $\rho\rightarrow0$ as
$n\rightarrow
\infty$, forcing all edge probabilities to decay at the same rate. Thus
it suffices to replace $\alpha$, $\beta$ or $\gamma$ by $\rho$ in
orders of magnitude; for example, the expected degree of nodes in
either class is $C_0n\rho$. We use ``$C_0$'' to denote a generic
positive constant.
All expectations are conditioned on $Z$; for notational convenience we
write $E[\cdot]$ instead of $E[\cdot|Z]$.

First we consider the eigenvalues and eigenvectors of $P$ without the
constraint of zero diagonals. If $\alpha\beta\neq\gamma^2$, then this
matrix (denoted by $P_B$) will have two eigenvalues with magnitude
$O(n\rho)$ and $n-2$ zero eigenvalues. Since $\|P-P_B\|=O(\rho)$, using
Weyl's inequality we see that the principal eigenvalues of $P$ are
$O(n\rho)$, whereas all other eigenvalues are $O(\rho)$.

Let $\mathbf{v}_i$ ($\lambda_i$) denote the $i$th eigenvector
(eigenvalue) of
matrix $P$. The ordering is in decreasing order of absolute value of
the eigenvalues.
We will denote the $i$th empirical eigenvector (eigenvalue) by $\hat
{\mathbf{v}}_i$ ($\hat{\lambda}_i$). $\mathbf{v}_1$, $\mathbf{v}_2$ are piecewise constant.

Now we will define the normalized counterparts of the above quantities.
Let $\tilde{A}:=D^{-1/2}AD^{-1/2}$, and also let $\tilde
{P}:=\mathcal{D}^{-1/2}P
\mathcal{D}^{-1/2}$, where $D$ and $\mathcal{D}$ are the diagonal
matrices of
degrees and expected degrees, respectively. We denote the first two
eigenvectors by $\mathbf{u}_1$ and $\mathbf{u}_2$, and the first two
eigenvalues by
$\nu_1$ and $\nu_2$. Similar to $\mathbf{v}_1$ and $\mathbf{v}_2$, $\mathbf{u}_1$ and $\mathbf
{u}_2$ also are
piecewise constant vectors. 
The empirical counterparts of the
eigenvectors and values are denoted by $\tilde{\mathbf{u}}_i$,
$\tilde{\nu
}_i$. One interesting fact about $\tilde{\mathbf{u}}_1$ is that the $i$th
entry is proportional to $\sqrt{d_i}$, where $d_i$ is the degree of
node $i$. However, one cannot explicitly obtain the form of $\hat
{\mathbf{u}}_1(i)$.

Among the many variants of spectral clustering, we consider the
algorithm used in \cite{rohechatterjiyu}. 
The\vspace*{1pt} idea is to compute an $n\times k$ matrix $\hat{Q}$ with the top $k$
eigenvectors of ${A}$ along its columns, and apply the
\texttt{kmeans} algorithm
on the rows of $\hat{Q}$. The \texttt{kmeans} algorithm searches over different
clusterings and returns a local optima of an objective function that
minimizes the squared Euclidean distance of points from their
respective cluster centers. The clusters are now identified as
estimates of the $k$ blocks.

Probabilistic bounds on misclassification errors of spectral clustering
under the stochastic blockmodel has been obtained in previous
work \cite{rohechatterjiyu,priebejasa2012}. However, upper bounds
cannot be used for comparing two algorithms. 
Instead, we define\vadjust{\goodbreak} a~simple clustering quality metric computable in
terms of an appropriately defined deviation of empirical eigenvectors
from their population counterparts, and we show that these are improved
by normalization.

\subsection{Quality metrics}
The quality metrics are defined as follows: the algorithm passes the
empirical eigenvectors to an oracle who knows the cluster memberships.
The oracle computes cluster centers $K_k:=\sum_{i\in C_k}\hat
{Q}_i/|C_k|$, for us $k\in\{1,2\}$. Let $d_{11}^2$ denote the mean
squared distance of points in $C_1$ from $K_1$, and let $d_{12}^2$
denote the mean squared distance of points in $C_1$ from $K_2$. From
now on we will denote by $\hat{d}_{11}^2$ and $\tilde{d}_{11}^2$ the
distances obtained from the unnormalized and normalized methods, respectively.

To be concrete, we can write $\hat{d}_{11}^2=\sum_{i\in C_1}\|\hat
{Q}_i-K_1\|^2/n\pi$. Similarly, define $\hat{d}_{12}^2$ as the mean
square distance of points in $C_1$ from $K_2$, that is, $\hat
{d}_{12}^2=\sum_{i\in C_1}\|\hat{Q}_i-K_2\|^2/n\pi$. One can
analogously define $\hat{d}_{22}^2$ and $\hat{d}_{21}^2$. We will use
the notation $d_{11}^2$ (or $d_{12}^2$) when we refer to the
corresponding quantities in general, that is, without any particular
reference to a specific method.

Although $\hat{d}_{11}^2$ seems like a simple average of squared
distances, it actually has useful information about the quality of clustering.
For definiteness, let us take the unnormalized case and examine points
in $C_1$. By stochastic equivalence, $\forall i\in C_1$, $\{\hat
{v}_1(i),\hat{v}_2(i)\}$ are\vspace*{1pt} identically distributed (albeit dependent)
random variables. Now $\hat{d}_{11}^2$ essentially is the trace of the
$2\times2$ sample variance matrix, and hence measures the variance of
these random variables.

Ideally a good clustering algorithm with or without normalization
should satisfy $d_{11}^2/d_{12}^2\stackrel{P}{\rightarrow}0$, but we
will show that this
ratio converges to zero at the same rate, with or without
normalization, in consistence with previous work
\cite{Oliveira2010,rohechatterjiyu} and \cite{priebejasa2012}.
Furthermore, we will show that $\tilde{d}_{12}^2/\hat
{d}_{12}^2\stackrel
{P}{\rightarrow}1$; that is, the two methods do not distinguish between
$d_{12}^2$.

Interestingly, our results also imply that $d_{11}^2$ increases as the
graphs become sparser; that is, $\rho$ decreases. Hence, if a method
can be shown to reduce the variance of points in a class by a constant
fraction, it would be preferable for sparse graphs. Indeed we show that
$\tilde{d}_{11}^2/\hat{d}_{11}^2$ converges to a constant which
is \textit{less than} 1 for a broad range of parameter settings of
$\alpha,\beta$
and $\gamma$. In the simple disconnected case with $\gamma=0$, this
constant is $1/4$.


Another advantage of $d_{11}^2$ is that it can be conveniently
expressed in terms of an appropriately defined deviation of empirical
eigenvectors from their population counterpart.
For any population and empirical eigenvector pair $\{\mathbf{v},\hat
{\mathbf{v}}\}$,\vspace*{1pt}
we consider the following orthogonal decomposition: $\mathbf{v}= c\hat
{\mathbf{v}
}+\mathbf{r}$, where $c:=\mathbf{v}^T\hat{\mathbf{v}}$. The norm of
residual $\mathbf{r}$ will
measure the deviation of $\hat{\mathbf{v}}$ from $\mathbf{v}$. The
deviation of
$\tilde{\mathbf{u}}$ from $\mathbf{u}$ can be measured similarly.

Since we are interested in two class blockmodels, we will mostly use
$\mathbf{r}_i, i\in\{1,2\}$ as the residual of the $i$th empirical eigenvector
from its population counterpart, and $c_{jj}:=\mathbf{v}_j^T\hat
{\mathbf{v}}_j$.
We denote by $v(C_1):=\sum_{i\in C_1} v(i)/n\pi$ the average of entries
of vector $\mathbf{v}$ restricted to class $C_1$. A key fact is that
$\mathbf{v}_1$,
$\mathbf{v}_2$ (or $\mathbf{u}_1$, $\mathbf{u}_2$) are both
piecewise constant:
\begin{eqnarray}
\label{eqd11} d_{11}^2 &=& 
\frac{1}{c_{11}^2} \biggl(\sum_{i\in C_1}
\frac{r_1(i)^2}{n\pi
}-r_1(C_1)^2 \biggr)+
\frac{1}{c_{22}^2} \biggl(\sum_{i\in C_1}
\frac
{r_2(i)^2}{n\pi}-r_2(C_1)^2 \biggr),
\\[-3pt]
\label{eqd12} d_{12}^2 &=& \frac{1}{n\pi}\sum
_{i\in C_1}\|\hat{Q}_i-K_2\|
^2=d_{11}^2+\| K_1-K_2
\|^2.
\end{eqnarray}
We will denote the distances obtained from $A$ by $\hat{d}_{\cdot
\cdot}$ and
from $\tilde{A}$ by $\tilde{d}_{\cdot\cdot}$.
Even though $K_i$ is defined in terms of $\mathbf{\hat{v}}_1$ and $\mathbf{\hat{v}}_2$,
we will abuse this notation somewhat to use the above expressions for calculating $\tilde{d}_{\cdot\cdot}$,
where $K_i$ will be defined identically in terms of $\tilde{\mathbf{u}}_1$ and $\tilde{\mathbf{u}}_2$.
For a wide regime of $(\alpha, \beta, \gamma)$, we prove that
$\tilde{d}_{11}^2$ is asymptotically a constant factor smaller and hence
better than $\hat{d}_{11}^2$.
First, using results from \cite{furedikomlos81} we will prove that
for $\gamma=0$, $\tilde{d}_{11}^2=1/4\hat{d}_{11}^2(1+o_P(1))$.
In this case, the result can be proven using existing results on Erd\H
{o}s--R\'{e}nyi
graphs~\cite{furedikomlos81} and a simple application of Taylor's
theorem. In order to generalize the result to $\gamma\neq0$, we would
need new convergence results for $A$ and $\tilde{A}$ generated
from a
stochastic blockmodel.
All results rely on the following assumption on $\rho$:

\begin{assumption}
\label{assumptionrho}
We assume $\log n/n\rho\rightarrow0$, as $n\rightarrow\infty$.
\end{assumption}

This assumption ascertains with high probability that the sequence of
growing graphs are not too sparse. The expected degree is $np=O(n\rho
)$, and this is the most commonly used regime where norm convergence of
matrices can be shown \cite
{Chung-radcliffe,Oliveira2010,chaudhuri-chung-2012}. Note that this is also
the sharp threshold for connectivity of Erd\H{o}s--R\'{e}nyi graphs
\cite{Bol98}. We
will now formally define the sparsity regime in which we derive our results.

\begin{defn}[(A semi-sparse stochastic blockmodel)]
\label{defSBM}
Define a stochastic blockmodel with parameters $\alpha$, $\beta$,
$\gamma$ and $Z$; see Definition~\ref{defSBM1}. Let $\alpha$, $\beta$
and $\gamma$ be deterministic quantities of the form $C_0\rho$. If
$\rho
$ satisfies Assumption~\ref{assumptionrho}, we call the stochastic
blockmodel $(\alpha,\beta,\gamma,Z)$ a semi-sparse stochastic blockmodel.
\end{defn}

The paper is organized as follows: we present the main results in
Section~\ref{secsomecomm-compare}. The proof of the simple $\gamma=0$
case is in Section~\ref{secnocomm}, whereas the expressions of $\hat
{d}_{11}^2$ and $\hat{d}_{12}^2$ in the general case appear in
Section~\ref{secmainres}.\vspace*{1pt} We derive the expressions of $\tilde{d}_{11}^2$
and $\tilde{d}_{12}^2$ in Section~\ref{secsomecomm-norm}.
Experiments on
simulated and real data appear in Section~\ref{secexp}.
The proofs of some accompanying lemmas and ancillary results are
omitted from the main manuscript for ease of exposition and are
deferred to the Supplement~\cite{suppnormalization}.

\subsection{Import of previous work}
\label{secims-previous}
By virtue of stochastic equivalence of points belonging to the same
class, eigenvectors of $P$ map the data to $k$ distinct points. This is
why consistency of spectral clustering is closely tied to consistency
properties of empirical eigenvalues and eigenvectors. We will show that
current theoretical work on eigenvector consistency does not
distinguish between the use of normalized or unnormalized $A$.


One of the earlier results on the consistency of spectral clustering
can be found in~\cite{luxburgConsistency08}, where weighted graphs
generated from a geometric generative model are considered.
While this is an important work, this does not apply to our random
network model.

For any symmetric adjacency matrix $A$ with independent entries, one
can use results on random matrix theory from Oliveira \cite
{Oliveira2010} to
show that the empirical eigenvectors of a semi-sparse stochastic
blockmodel converge to their population counterpart at the same rate
with or without normalization.
If $\mathbf{p}\dvtx[n]^2\rightarrow[0,1]$ denotes the probability function
$P(A_{ij}=1)=1-P(A_{ij}=0)=\mathbf{p}(i,j)$, and $d_{\mathbf{p}}$
denotes the expected degree, then:

\begin{theorem}[(Theorem~3.1 of~\cite{Oliveira2010})]
\label{thmoliveira}
For any constant $c > 0$, there exists another constant $C = C(c) > 0$,
independent of $n$ or $\mathbf{p}$, such that the following holds.
Let $d:=\min_{i\in[n]}d_{\mathbf{p}}(i)$, $\Delta:=\max_{i\in
[n]}d_{\mathbf{p}}(i)$. If $\Delta>C\log n$, then for all $n^{-c}\leq
\delta\leq1/2$,
\[
P \bigl(\|A-P\|\leq4\sqrt{\Delta\log(n/\delta)} \bigr)\geq 1-\delta.
\]
Moreover, if $d\geq C\log n$, then for the same range of $\delta$,
\[
P \bigl(\|\tilde{A}-\tilde{P}\|\leq14\sqrt{\log(4n/\delta )/d} \bigr)
\geq1-\delta.
\]
\end{theorem}

Let $\Pi_{a,b}(A)$ denote the orthogonal projector onto the space
spanned by the eigenvectors of $A$ corresponding to eigenvalues in
$[a,b]$. A simple consequence of Theorem~\ref{thmoliveira} is that for
suitably separated population eigenvalues, the operator norm of the
difference of the eigenspaces also converges to zero.

\begin{cor}[(Corollary~3.2 of~\cite{Oliveira2010})]
Given some $x>0$, let $N_x(P)$ be the set of all pairs $a<b$ such that
$a+x<b-x$, and $P$ has no eigenvalues in $(a-x,a+x)\cup(b-x,b+x)$.
Then for $x>4\sqrt{\Delta\log(n/\delta)}$,
%
\begin{eqnarray}
&&\|A-P\|\leq4\sqrt{\Delta\log(n/\delta)}\nonumber\\
&& \qquad\Longrightarrow\quad \forall(a,b)\in N_x(P),\nonumber\\
\eqntext{\displaystyle\bigl\|\Pi
_{a,b}(A)-\Pi_{a,b}(P)\bigr\|\leq \biggl(\frac{4(b-a+2x)}{\pi(x^2-x\sqrt{\Delta\log(n/\delta
)})} \biggr)
\sqrt{\Delta \log(n/\delta)}.} 
\end{eqnarray}
Similarly define $N_x(\tilde{P})$. Then for $x>14\sqrt{\log
(4n/\delta)/d}$,
%
\begin{eqnarray}
&& \bigl\|\tilde{A}-\tilde{P}\bigr\|\leq14\sqrt{\frac{\log
(4n/\delta)}{d}}\nonumber\\
&& \qquad\Longrightarrow\quad\forall(a,b)\in N_x(\tilde {P}),\nonumber\\
\eqntext{\displaystyle\bigl\|\Pi_{a,b}(\tilde{A})-\Pi_{a,b}(\tilde{P})\bigr\| \leq
\biggl(\frac{4(b-a+2x)}{\pi(x^2-x\sqrt{\Delta\log(n/\delta
)})} \biggr)\sqrt{\log(n/\delta)/d}.}
\end{eqnarray}
In particular the right-hand sides of the above equations hold with
probability $\geq1-\delta$ for any $n^{-c}<\delta<1/2$.
\end{cor}

A straightforward application of this corollary yields that spectral
clustering for a stochastic blockmodel with $A$ and $\tilde{A}$
lead to
$O_P(\sqrt{\log n/n\rho})$ convergence of empirical eigenvectors to
their population counterparts. Further analysis shows that the fraction
of misclassified nodes go to zero at the same rate for $A$ and
$\tilde{A}$.
We defer the proof to Section~B of the Supplement \cite{suppnormalization}.

\begin{cor}
\label{coroliveira-sbm}
Let $A$ be generated from a semi-sparse stochastic blockmodel
(Definition~\ref{defSBM}) with $\gamma>0$ and $\alpha\beta\neq
\gamma
^2$. Then, for $i\in\{1,2\}$, $\|\mathbf{v}_i\mathbf{v}_i^T-\hat
{\mathbf{v}}_i\hat{\mathbf{v}}_i^T\|
=O_P(\sqrt{\log n/n\rho})$. Furthermore the fraction of misclassified
nodes can be bounded by $O_P(\log n/n\rho)$ for both methods.
\end{cor}



Spectral clustering with $\tilde{A}$ derived from a stochastic blockmodel
with growing number of blocks has been shown to be asymptotically
consistent~\cite{rohechatterjiyu}. Further, the fraction of
mis-clustered nodes is shown to converge to zero under general
conditions. These results are extended to show that spectral clustering
on unnormalized $A$ also enjoys similar asymptotic properties~\cite
{priebejasa2012}. Sussman et~al. \cite{priebejasa2012} also give an
example of
parameter setting for a stochastic blockmodel where spectral clustering
using unnormalized $A$ outperforms that using $\tilde{A}$. We, however,
demonstrate using theory and experiments that this is only a partial
picture, and there is a large regime of parameters where normalization
indeed improves performance.

For ease of exposition, we list the different variables and their
orders of magnitude in Table~\ref{tablenotation}.
For deterministic quantities $x_n$ and $c_n$, $x_n\asymp c_n$, denotes
that $x_n/c_n$ converges to some constant as $n\rightarrow\infty$. For
two random variables $X_n$ and $Y_n$, we use $X_n\sim Y_n$ to denote
$X_n=Y_n(1+o_P(1))$. For the scope of this paper $\|\cdot\|$ denotes the
$L_2$ norm, unless otherwise specified.
%
\begin{table}
\tabcolsep=0pt
\tablewidth=\textwidth
\caption{Table of notation}
\label{tablenotation}
{\fontsize{8.4pt}{10.4pt}\selectfont\begin{tabular*}{\tablewidth}{@{\extracolsep{\fill}}llll@{}}\hline
$\rho$& Edge probability & $I$ & The $n\times n$ identity matrix\\
$n$&Number of nodes in the network&$Z$ & $n\times2$ binary matrix of
class memberships\\
$C_i$&The $i$th group, $i\in\{1,2\}$&$\pi$& $|C_1|/n$\\
$D$&Diagonal matrix of degrees&$\mathcal{D}$&Diagonal matrix of
expected degrees,\\
&&& conditioned on $Z$\\
$A$&Adjacency matrix &$\tilde{A}$& $D^{-1/2}AD^{-1/2}$\\
$P$&$E[A|Z]$&$\tilde{P}$&$\mathcal{D}^{-1/2}P\mathcal
{D}^{-1/2}$\\
$\alpha$& $P[A_{ij}=1|i\in C_1,j\in C_1]\asymp\rho$&$\mu
_1$&$E[D_{ii}/n| i\in C_1] =\pi\alpha+(1-\pi)\gamma-\alpha/n\asymp
\rho
$\\
$\beta$& $P[A_{ij}=1|i\in C_2,j\in C_2]\asymp\rho$&$\mu
_2$&$E[D_{ii}/n| i\in C_2] =\pi\gamma+(1-\pi)\beta-\beta/n\asymp
\rho$\\
$\gamma$& $P[A_{ij}=1|i\in C_1,j\in C_2]\asymp\rho$&$\mu$&$\sum_{i}\mathcal{D}_{ii}/n^2=\pi\mu_1+(1-\pi)\mu_2\asymp\rho$\\
$d_i$&$D_{ii},i\in\{1,\dots n\}= O_P(n\rho)$&$\bar{d}_i$&$\sum_{j} [A_{ij}-E[A_{ij}|Z]]=O_P(\sqrt{n\rho})$ \\
$\bar{d}_i^{(1)}$&$\sum_{j\in C_1} [A_{ij}-E[A_{ij}|Z]]=O_P(\sqrt
{n\rho})$
&$\bar{d}_i^{(2)}$&$\sum_{j\in C_2} [A_{ij}-E[A_{ij}|Z]]=O_P(\sqrt
{n\rho})$
\\
$E_1$&$\sum_{i\in C_1}d_i$&$E_2$&$\sum_{i\in C_2}d_i$\\
$E$&$\sum_i d_i$&$x(C_1)$&The average of $\mathbf{x}$ restricted to $C_1$,
that is,\\
&&& $\sum_{i\in C_1}\mathbf{x}(i)$\\
$\lambda_i$ & $i$th largest eigenvalue of $P$ in magnitude & $\nu_i$
&$i$th largest eigenvalue of $\tilde{P}$ in magnitude\\
&$\asymp n\rho$, for $i\in\{1,2\}$&&$\asymp1$, for $i\in\{1,2\}$\\
$\mathbf{v}_i$&$i$th eigenvector of $P$&$\mathbf{u}_i$& $i$th
eigenvector of $\tilde{P}$\\
$x_k$&$\mathbf{v}_k(i)  (k\in\{1,2\}, i\in C_1)\asymp1/\sqrt
{n}$&$\tilde{x}_k$&$\mathbf{u}_k(i)  (k\in\{1,2\}, i\in
C_1)\asymp1/\sqrt{n}$\\
$y_k$&$\mathbf{v}_k(i) (k\in\{1,2\}, i\in C_2)\asymp1/\sqrt
{n}$&$\tilde{y}_k$&$\mathbf{u}_k(i) (k\in\{1,2\}, i\in
C_2)\asymp1/\sqrt{n}$\\
$\hat{\lambda}_i$&$i$th largest eigenvalue of $A$ in magnitude;
&$\tilde{\nu }_i$&$i$th largest eigenvalue of $\tilde{A}$ in
magnitude\\
$\hat{\mathbf{v}}_i$&$i$th eigenvector of $A$&$\tilde{\mathbf
{u}}_i$&$i$th eigenvector of
$\tilde{A}$\\
$K_1$&$\sum_{j\in C_1} \hat{Q}_j/n\pi$&$K_2$&$\sum_{j\in
C_2} \hat{Q}_j/n\pi$\\
$\hat{Q}$& $n\times2$ matrix of top two empirical&$Q$& The population
variant of $\hat{Q}$\\
& eigenvectors (of $A$) along the columns&&\\
$\hat{d}_{k\ell}^2$&$\sum_{i\in C_k}\|\hat{Q}_i-K_\ell\|^2/n\pi$
&$\tilde{d}_{k\ell}^2$& Variant of $\hat{d}_{k\ell}^2$ using
eigenvectors of $\tilde{A}$\\[2pt]
$C$&$\hat{Q}^T Q$&$c_{ij}$ & $C_{ij}:=\mathbf{v}_i^T\hat{\mathbf
{v}}_j$\\[2pt]
$\hat{\mathbf{r}}_i$&$\mathbf{v}_i-(\mathbf{v}_i^T\hat{\mathbf
{v}}_i)\hat{\mathbf{v}}_i, i\in\{1,2\}
$&$\tilde{\mathbf{r}}_i$&$\mathbf{u}_i-(\mathbf
{u}_i^T\tilde{\mathbf{u}_i})\tilde{\mathbf{u}_i}, i\in\{
1,2\}$\\
\hline
\end{tabular*}}
\end{table}
\section{Main results}
\label{secsomecomm-compare}
For the general case we derive the following asymptotic expressions of
$d_{11}^2$ and $d_{12}^2$. We recall that $d_{11}^2$ measures the
variance of points in class one under the spectral representation,
whereas $d_{12}^2$ basically measures the distance between the class
centers, which can also be thought of as bias.\vadjust{\goodbreak} We will show that
normalizing $A$ asymptotically reduces the variance without affecting
the bias.
The proofs can be found in Sections~\ref{secmainres} and~\ref
{secsomecomm-norm}.
%
\begin{thm}
\label{thmlocommdistunnorm}
Let $A$ be the adjacency matrix generated from a semi-sparse stochastic
blockmodel $(\alpha,\beta,\gamma,Z)$ where $\gamma>0$ and $\alpha
\beta
\neq\gamma^2$. 
We define $\lambda_i$, $x_i$, $y_i$, for $i\in\{1,2\}$ and $\pi$ as in
Table~\ref{tablenotation}:
%
\begin{eqnarray}
\label{eqlowcomm-final-d11-bound} 
\hspace*{6pt}\quad \hat{d}_{11}^2 &\sim& \biggl[ \biggl(\frac{x_1^2}{\lambda
_1^2}+
\frac {x_2^2}{\lambda_2^2} \biggr)n\pi\alpha(1-\alpha)+ \biggl(\frac{y_1^2}{\lambda _1^2}+
\frac{y_2^2}{\lambda_2^2} \biggr)n(1-\pi)\gamma(1-\gamma) \biggr],
\\
\label{eqlowcomm-final-d12-bound} \hat{d}_{12}^2 &\sim&
1/n\pi(1-\pi).
\end{eqnarray}
%
\end{thm}
%
\begin{thm}\label{thmd11-sq-norm}
Let $A$ be the adjacency matrix generated from a semi-sparse stochastic
blockmodel $(\alpha,\beta,\gamma,Z)$ where $\gamma>0$ and $\alpha
\beta
\neq\gamma^2$.
We define $\mu_1$, $\mu_2$, $\nu_2$ and $\pi$ as in\vspace*{-2pt}
Table~\ref{tablenotation}:
%
\begin{eqnarray}
\tilde{d}_{11}^2 &\sim& \biggl[
\frac{n\pi\alpha(1-\alpha)}{n^3\pi\mu_1^2} \biggl(\frac
{1}{4}+\frac{(1-\pi)\gamma}{\mu_1\nu_2^2} \biggr)
\nonumber
\\[-9pt]
\label{eqsomecomm-norm-d11SQ}\\[-9pt]
\nonumber
&&\hspace*{-6pt}\quad{}+
\frac{n(1-\pi)\gamma(1-\gamma)}{n^3\mu_1^2} \biggl(\frac{1}{4\pi
}+\frac{\pi\alpha}{(1-\pi)\mu_2\nu_2^2} \biggr)\biggr],
\\[-3pt]
\label{eqsomecomm-norm-d12SQ} \tilde{d}_{12}^2
&\sim& \frac{1}{n\pi(1-\pi)}.\vspace*{-3pt}
\end{eqnarray}
\end{thm}

Before explaining the above theorems, we present a special case for clarity.
\begin{remark*}[(A special case)]
When $\gamma=0$, we have $\lambda_1=n\mu_1$, and $x_1=1/\sqrt{n\pi}$
and $y_1=0$, which immediately shows that normalization shrinks the
variance of the spectral embedding within a class ($d_{11}^2$) by a
factor of four. This is the completely associative case. Now consider
the completely dissociative case, that is, $\gamma=0$. It is easy to
see that then $\lambda_1=-\lambda_2=n\sqrt{\pi(1-\pi)}\gamma$, and
$y_1=-y_2=1/\sqrt{2n(1-\pi)}$. Substituting these values into the
distance formulas again shows that normalization shrinks $d_{11}^2$ by
a factor of four in the completely dissociative case.

We call the completely associative case the zero communication case,
which can be thought of as two disconnected Erd\H{o}s--R\'{e}nyi graphs. Under
Assumption~\ref{assumptionrho} each of the smaller graphs will be
connected with probability tending to one. von Luxburg \cite{LuxburgTutorial07}
already established that spectral clustering achieves perfect
classification in this
scenario. We merely present this simple setting because the ideas and
proof techniques used for this case will be carried over to the general
case with $\gamma\neq0$. In particular, our results indicate that in
the general case ($\alpha,\gamma>0$), for parameter regimes close to
the completely associative or completely dissociative models, the
normalized method has a clear\vspace*{-2pt} edge.
\end{remark*}

\begin{cor}
\label{corerdos-norm-4}
Let $A$ be the adjacency matrix generated from a semi-sparse stochastic
blockmodel $(\alpha,\beta,\gamma,Z)$ (see Definition~\ref{defSBM})
where $\gamma=0$ and $\alpha\beta\neq\gamma^2$. 
We\vspace*{-3pt} have
%
\begin{eqnarray}
\label{eqd11-result-nocomm} \frac{\hat{d}_{11}^2}{\tilde{d}_{11}^2}&\sim& 4,
\\[-2pt]
\label{eqd12-result-nocomm} \frac{\hat{d}_{12}^2}{\tilde{d}_{12}^2} &\sim& 1,
\\[-2pt]
\label{eqd11-d12-ratio-nocomm-unnorm} \frac{\hat{d}_{11}^2}{\hat{d}_{12}^2}&=& O_P \biggl(\frac{1}{n\rho
}
\biggr),
\\[-2pt]
\label{eqd11-d12-ratio-nocomm-norm} \frac{\tilde{d}_{11}^2}{\tilde{d}_{12}^2}&=& O_P \biggl(\frac
{1}{n\rho}
\biggr).
\end{eqnarray}

The same holds for normalized and unnormalized versions of $d_{22}^2$
and $d_{21}^2$.
\end{cor}

While both of $d_{11}^2$ and $d_{12}^2$ (derived the unnormalized and
normalized methods) are approaching zero in probability,
$d_{11}^2/d_{12}^2$ is $C_0/n\rho$ for both the normalized and
unnormalized cases. In our regime of $\rho$ this translates to perfect
classification as $n\rightarrow\infty$. This is not unexpected because
existing literature has established that spectral clustering with both
$A$ and $\tilde{A}$ are consistent in the semi-sparse regime.
Also, $\tilde{d}_{12}^2/\hat
{d}_{12}^2$ approaches one; thus if the limiting ratio $\tilde
{d}_{11}^2/\hat{d}_{11}^2$ is smaller (larger) than one, then there is
some indication that the normalized (unnormalized) method is to be preferred.

For simplicity, we consider a stochastic blockmodel with two equal
sized classes and
$\beta=\alpha$. We will now show that for this simple model, in the
semi-sparse regime, our quality metric indicates that normalization
would always improve performance. In the dense regime, that is, when
degree grows linearly with $n$, there are parts of the parameter space
where our quality metric prefers the unnormalized method. However, the
network is so dense that the two class centers are well separated,
leading to equally good performance of both\vspace*{1pt} methods.

\begin{figure}[t]
\begin{tabular}{@{}c@{\hspace*{6pt}}c@{}}

\includegraphics{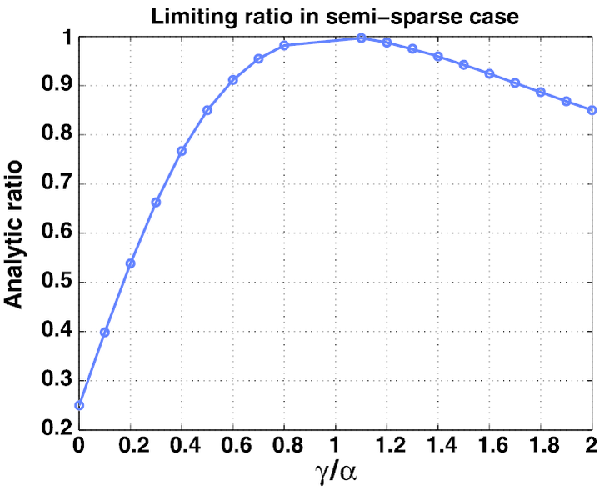}
 & \includegraphics{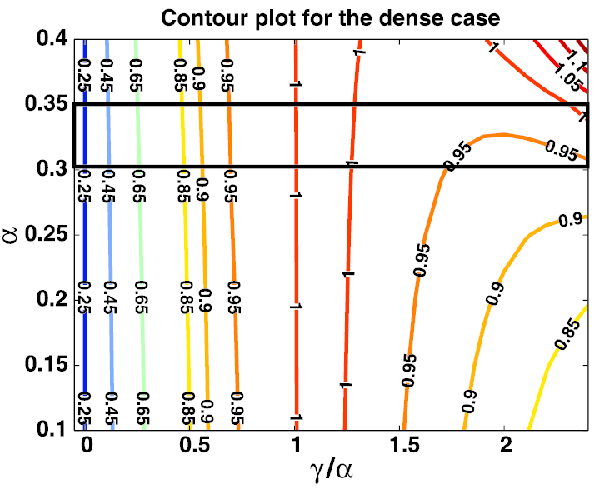} \\
\footnotesize{(A)} & \footnotesize{(B)}\\[6pt]
\multicolumn{2}{c}{
\includegraphics{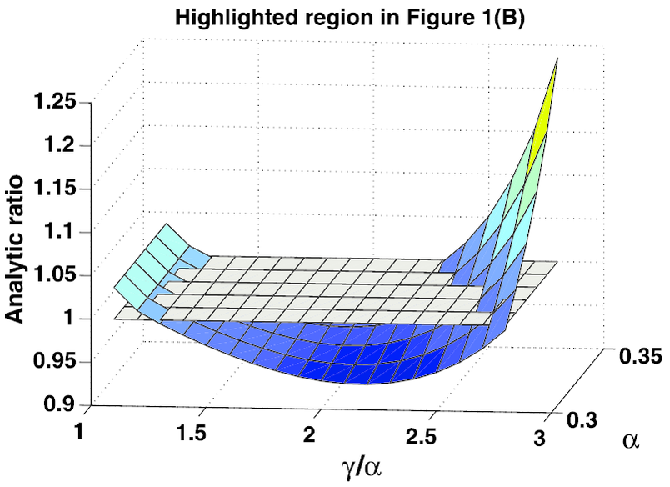}
}\\
\multicolumn{2}{c}{\footnotesize{(C)}}
\end{tabular}
\caption{Simple blockmodel with two equal sized
classes and $\alpha=\beta$: \textup{(A)} Limiting ratio of $\tilde
{d}_{11}^2/\hat
{d}_{11}^2$ in the semi-sparse case.
\textup{(B)} Contour plot for the ratio in the dense case. The rectangular area
consists of parameters settings leading to a ratio bigger than one. This is highlighted in \textup{(C)},
which shows the surface plot for $\tilde{d}_{11}^2/\hat{d}_{11}^2$ along the
$Z$ axis in
the regime where ratio is close to or larger than one. $Y$ axis has
varying $\alpha$, $X$~axis has varying $\gamma/\alpha$. For reference we
also plot the plane $Z=1$.}\vspace*{-4pt}\label{figanalytic}
\end{figure}

For this simple model, the limiting $\tilde{d}_{11}^2/\hat
{d}_{11}^2$ ratio
has a concise form in the semi-sparse case, which is presented in
Corollary~\ref{corsimple}, and plotted in Figure~\ref{figanalytic}(A). Figure~\ref{figanalytic}(B) shows the contour plot
of the limiting ratio in the dense case. We also highlight the
parameter regime where the ratio is close to or larger than one.
Finally Figure~\ref{figanalytic}(C) focuses on this highlighted area.

\begin{cor}
\label{corsimple}
Let $A$ be the adjacency matrix generated from the stochastic
blockmodel $(\alpha,\alpha,\gamma,Z)$ where $\gamma=x\alpha>0$ and
$\pi=1/2$. When $\rho\rightarrow0$, we have the following limit, which
is always smaller than one:
%
\begin{equation}
\label{eqlimitsparse} \frac{\tilde{d}_{11}^2}{\hat{d}_{11}^2} \sim\frac{1}{4}+\frac
{3}{2}
\frac{x}{1+x^2}.
\end{equation}
On the other hand, when $\rho$ is a constant w.r.t. $n$, the above ratio
is smaller than one, unless $x\geq1$ or $\alpha\geq1/2$. The
universal upper bound is 1.31.
\end{cor}

\begin{remark*}
Here we summarize the result in the above corollary.
\begin{longlist}[(3)]
\item[(1)] In the semi-sparse regime [Figure~\ref{figanalytic}(A)],
the limiting ratio is always less than one, thus favoring the
normalized method.
\item[(2)] In the dense case [Figure~\ref{figanalytic}(B) and (C)],
where $\rho$ is a constant w.r.t. $n$, this ratio can be larger than one
when $x\geq1$ or $\alpha\geq1/2$, with an upper bound of 1.31. The
upper bound is achieved for large $\alpha$, $\gamma$ pairs, for
example, $\alpha=1/3,\gamma=1$ and $\alpha=1,\gamma=0.24$. In this dense
regime, both methods perform equally well on any reasonably sized
network. Using simulations on small networks (twenty nodes), we found
that in terms of misclassification error, the methods perform comparably.
\item[(3)] Because of the inherent symmetry of the simple model, for
$y:=\alpha/\gamma$, the ratio $\tilde{d}_{11}^2/\hat
{d}_{11}^2\rightarrow
1/4+3y/2(1+y^2)$, in the semi-sparse regime. This again shows that
normalization provides a clear edge close to the completely
associative
($\gamma=0$) or completely dissociative ($\alpha=0$) cases.
\end{longlist}
\end{remark*}
We want to point out that for the simulated experiment with $\alpha
=0.42,\beta=0.50,\gamma=0.42$ (and $\pi=0.60$), the unnormalized method
performs better than the normalized method
in
Sussman et~al. \cite{priebejasa2012}.
In this case the analytic ratio also is larger than one, and the graph
is very close to an Erd\H{o}s--R\'{e}nyi graph.
\subsection{A shortcoming of asymptotic analysis}
While Corollary~\ref{corsimple} suggests that normalization always
reduces within class variance in the semi-sparse degree regime, there
is one caveat to this asymptotic result. In the semi-sparse regime, the
network is connected with high probability as $n\rightarrow\infty$.
However, finite sparse networks may have disconnected components
consisting of a few nodes. In such scenarios, by construction the
normalized method assigns eigenvalue one to eigenvectors with support
on nodes in each of the connected components. As a result, the leading
eigenvectors are uninformative, leading to poor performance. The
unnormalized method, however, does not suffer from this problem and has
the informative eigenvectors as the leading eigenvectors, albeit with a
high variance. We get around this problem by removing small degree
nodes and then working on the largest connected component. We also
point out this problem in the discussion section.

\subsection{Accuracy of the analytic ratio}
Finally, we also use simulations to see how accurate the analytic ratio
is. For $n=1000$, we vary $\alpha\in[0.4, 0.6]$, $\beta\in[0.5, 0.9]$
and $\gamma/\alpha$ between zero and two such that $\forall\alpha$,
$\gamma\leq1$, and $\gamma^2\neq\alpha\beta$. We note that the ratio
increases for large $(\alpha,\gamma)$ pairs.\vspace*{1pt} The mean, median and
maximum absolute relative error for $\tilde{d}_{11}^2/\hat
{d}_{11}^2$ ($\tilde{d}_{12}^2/\hat{d}_{12}^2$) from\vspace*{1pt} their
analytic counterparts is 0.02,
0.02 and 0.1 (0.001, 0.001 and 0.03), respectively.
In both cases the maximum happens for the $\{\alpha,\beta,\gamma\}$
combination where $|\alpha\beta-\gamma^2|$ is the smallest, leading to
most instability. Since all our $o_P(1)$ terms are $O_P(1/\sqrt{n\rho
})$, for this experiment these errors are indeed justifiable.

\section{The zero communication case}
\label{secnocomm}
We will now present our result for two class blockmodels (see
Definition~\ref{defSBM}) with $\gamma=0$.
We will heavily use the following orthogonal decomposition of the
population eigenvectors:
\[
\mathbf{v}_k:=c_{kk}\hat{\mathbf{v}}_k+
\mathbf{r}_k \qquad\mbox {for $k\in\{1,2\}$, where
$c_{kk}=\mathbf{v}_k^T\hat{
\mathbf{v}}_k$}.
\]
Since $\gamma=0$, $A$ can be thought of as two disconnected Erd\H{o}s--R\'{e}nyi graphs
of size $n\pi$ and $n(1-\pi)$ (let the two adjacency matrices be
denoted by $A_1$ and $A_2$, resp.). 
We assume WLOG $\pi\alpha>(1-\pi)\beta$ so that $\lambda_1=n\pi
\alpha
+O(\rho)$ and $\lambda_2=n(1-\pi)\beta+O(\rho)$. We also assume that
rows and columns of $A$ are permuted so that the first $n\pi$ entries
are from $C_1$. (We will not use this in our proofs; it only helps the
exposition.)

F\"{u}redi and Koml\'{o}s
\cite{furedikomlos81} show that for $i\in\{1,2\}$, $\hat{\lambda
}_i=\lambda_i+O_P(1)$ and $\max_{i>2}|\lambda_i|=O_P(\sqrt{n\rho})$.
Hence for large $n$, the second largest eigenvalue will come from
$A_2$, and will have zeros along the first class, similar to the
second population eigenvector. Thus $\hat{r}_1(i)=0$ for $i\in C_2$,
and vice versa.

Further, some algebra reveals that $K_1=\{c_{11}/\sqrt{n\pi
},0\}
$ and $K_2=\{0,c_{22}/ \break \sqrt{n(1-\pi)}\}$. Computing $\hat
{d}_{11}^2$ or $\tilde{d}_{11}^2$ requires one to compute the norm and
average of $\hat{\mathbf{r}}_k$ and $\tilde{\mathbf{r}}_k$
restricted to $C_1$; see equation~(\ref{eqd11}).
For $\gamma=0$, this reduces to examining $\hat{\mathbf{r}}$ and
$\tilde{\mathbf{r}}$ for
two Erd\H{o}s--R\'{e}nyi graphs.

Let us consider an Erd\H{o}s--R\'{e}nyi graph $G_{n,p}$. Since
self-loops are
prohibited, the conditional expectation matrix $P$ is simply $p(\mathbf
{1}\mathbf{1}
^T-I)$, which has $n$ eigenvalues, the largest of which is $\lambda
:=(n-1)p$, and the rest are all $-p$. We denote by $d_i$ the degree of
node $i$, and $\bar{d}_i:=d_i-(n-1)p$.

Let $\lambda$, $\mathbf{v}$ $(\nu, \mathbf{u})$ be respectively the
principal eigenvalue and eigenvector pair of $P$ ($\tilde{P}$), whose
empirical counterparts are given by $\hat{\lambda}$ and $\hat{\mathbf{v}}$
($\hat{\nu}$ and $\tilde{\mathbf{u}}$) respectively. In this simple case, $\mathbf{v}$ and $\mathbf{u}$ are the same.
We require that all eigenvectors are unit-length. We
denote by $\langle x_i\rangle$ the a $n$
length vector with the $i$th entry equaling~$x_i$.
We note that $\mathbf{v}=\langle1/\sqrt{n}\rangle$, and $\tilde
{\mathbf{v}}= \langle \sqrt{d_i/\sum_j d_j}
 \rangle$. Let $\hat{c}:=\hat{\mathbf{v}}^T\mathbf{v}$ and
$\tilde{c}:=\tilde{\mathbf{u}}^T\mathbf{u}$.
We will prove that $\|\hat{\mathbf{r}}\|^2\sim4\|\tilde{\mathbf
{r}}\|^2$, which will help us prove
Corollary~\ref{corerdos-norm-4}.

Before proceeding with the result, for ease of exposition we recall the
orders of magnitudes of some random variables used in the proof. Let
$E$ denote $\sum_i d_i$.
We have $\sum_i\bar{d}_i=O_P(n\sqrt{\rho})$ [this is simply twice
the sum of
$n\choose2$ centered $\operatorname{Bernoulli}(p)$ variables and $\sum_i\bar{d}_i^2 =
n^2p(1+o_P(1))$]. The later result can be obtained by showing that the
expectation is $n(n-1)p$, and the standard deviation is of a smaller
order. A detailed proof can be found in~\cite{furedikomlos81}.

\begin{lem}
\label{lemr-unnorm}
Write the first population eigenvector $\mathbf{v}$ of an Erd\H
{o}s--R\'{e}nyi $(n,p)$ graph
adjacency matrix $A$ as $\mathbf{v}= \hat{c}\hat{\mathbf{v}}+\hat
{\mathbf{r}}$. If
$p=C_0\rho$ satisfies Assumption~\ref{assumptionrho},
we have
\[
\|\hat{\mathbf{r}}\|^2\sim\frac{1}{((n-1)p)^2} \frac{\sum_i\bar
{d}_i^2}{n}.
\]
\end{lem}

\begin{pf}
Before delving into the proof, we state the main result from~\cite
{furedikomlos81}. For an Erd\H{o}s--R\'{e}nyi graph, $\hat{\lambda
}_1=\frac{\mathbf{1}^T A \mathbf{1}
}{n}+ (1-p) + O_P(1/\sqrt{n})$.
Since $\frac{\mathbf{1}^T A \mathbf{1}}{n}-(n-1)p = O_P (\sqrt
{p(1-p)} )$, we have
$\hat{\lambda}_1-(n-1)p = O_P(1)$. As the explicit form of~$\hat
{\mathbf{v}}$
is not known, the following step is used to compute the norm of $\hat
{\mathbf{r}}$:
%
\begin{equation}
\label{eq-r-expr} (A-\hat{\lambda}_1I)\hat{\mathbf{r}}= \bigl
\langle{(d_i-\hat {\lambda}_1)/\sqrt{n}} \bigr\rangle.
\end{equation}
The proof is straightforward. First we see that
\[
A\mathbf{v}-\hat{\lambda}_1\mathbf{v}=A(
\hat{c}\hat{\mathbf {v}}+\hat{\mathbf{r}})-\hat{\lambda}_1\mathbf{v}
=(A-\hat{\lambda}_1I)\hat{\mathbf{r}}.
\]
Using $\mathbf{v}= \langle{1/\sqrt{n}}  \rangle$,
$A\mathbf{v}-\hat{\lambda}_1\mathbf{v}= \langle{(d_i-\hat
{\lambda}_1)/\sqrt{n}}  \rangle$, thus proving equation~(\ref
{eq-r-expr}).
Now equation~(\ref{eq-r-expr}) and standard norm-inequalities yield $\|
A-\hat{\lambda}_1I\|\leq(\hat{\lambda}_1+\max(\hat{\lambda
}_2,|\hat
{\lambda}_n|))$, where $\hat{\lambda}_i$ is the $i$th largest
eigenvalue of $A$.

Now, using results from~\cite{feigeofek} we have $\max(\hat{\lambda
}_2,|\hat{\lambda}_n|)=O_P(\sqrt{np})$, and hence $\|A-\hat{\lambda
}_1I\|\sim np$.
Interestingly, note that $\hat{\mathbf{r}}\perp\hat{\mathbf{v}}$,
and hence $\|A\hat{\mathbf{r}}\|/\|\hat{\mathbf{r}}
\| =\break O_P(\sqrt{np})$. Hence $\|(A-\hat{\lambda}_1I)\hat{\mathbf
{r}}\|\geq\hat{\lambda
}_1(1+o_P(1))\|\hat{\mathbf{r}}\|$. Combining this with the former
upper bound, we
have
\[
\bigl\|(A-\hat{\lambda}_1I)\hat{\mathbf{r}}\bigr\|\sim\hat{
\lambda}_1\| \hat{\mathbf{r}}\|.
\]
Since $\bar{d}_i/\sqrt{n}=O_P(\sqrt{p(1-p)})$ and $E[d_i]=(n-1)p$,
we have
\[
\sum_i \frac{(d_i-\hat{\lambda}_1)^2}{n} 
=
\sum_i\frac{\bar{d}_i^2}{n}+\bigl(\hat{
\lambda}_1-(n-1)p\bigr)^2-2\bigl(\hat {\lambda
}_1-(n-1)p\bigr)\frac{\sum_i \bar{d}_i}{n} \sim\sum
_i\frac{\bar{d}_i^2}{n}.
\]
The last step is true because $\sum_i\frac{\bar{d}_i^2}{n}=O_P(np(1-p))$,
whereas both $\hat{\lambda}_1-(n-1)p$ and $\sum_i\bar{d}_i/n$ are $O_P(1)$.
Simple application of the Cauchy--Schwarz inequality shows that the
cross term is also $O_P(1)$.
Now we have
%
\begin{equation}
\label{eq-unnorm-lb} \hat{\mathbf{r}}^T\hat{\mathbf{r}}=\frac{1}{\hat{\lambda
}_1^2}
\sum_i \frac{(d_i-\hat{\lambda
}_1)^2}{n}\sim\frac{1}{((n-1)p)^2}
\frac{\sum_i\bar{d}_i^2}{n}.
\end{equation}
\upqed\end{pf}
%
Since the form of $\tilde{\mathbf{u}}$ is known, $\tilde
{\mathbf{r}}^T\tilde{\mathbf{r}}$ can be
obtained by using element-wise Taylor expansion. The only complication
arises because we often approximate the norm of a length $n$ vector by
the norm of its first or second order Taylor expansion, where $n$ is
growing to infinity. Hence we present the following helping lemma,
where we formalize sufficient conditions for neglecting lower order
terms in such an expansion.

\begin{lem}
\label{lemnorm-centered-taylor}
Consider length $n$ vector $\mathbf{x}_n:=c_n+\mathbf
{x}_n^{1}+\mathbf
{R}_n$ where $c_n$ is a vector of constants $c$.
If both $\|\mathbf{R}_n\|=o_P(\|\mathbf{x}_n^{1}\|)$ and $|\sum_i
x_n^{1}(i)/n|=o_P(\|\mathbf{x}_n^{1}\|/\sqrt{n})$,
as $n\rightarrow\infty$,
$\sum_i (x_n(i)-\sum_i{x_n(i)}/n)^2\sim\|\mathbf{x}_n^{1}\|^2$.%
\end{lem}

The following lemma has the asymptotic form of $\|\tilde{\mathbf
{r}}\|^2$.

\begin{lem}\label{lemr-norm}
Write the first population eigenvector $\mathbf{u}$ of an Erd\H
{o}s--R\'{e}nyi $(n,p)$ graph
normalized adjacency matrix $\tilde{A}$ as $\mathbf{u}=
\tilde{c}\tilde{\mathbf{u}}+\tilde{\mathbf{r}}$. If
$p=O(\rho)$ satisfies Assumption~\ref{assumptionrho},
\[
\|\tilde{\mathbf{r}}\|^2\sim\frac{1}{4n(n-1)p^2}\sum
_i \frac
{\bar{d}_i^2}{n}.
\]
\end{lem}

\begin{pf*}{Proof Sketch}
We will use the fact that $\|\tilde{\mathbf{r}}\|^2=1-\tilde
{c}^2=\sum_i (\tilde{u}_i -
\sum_i\tilde{u}_i/n)^2$. Since one can explicitly obtain the
expression of
$u_i$, the basic idea is to use Taylor approximation term by term to
obtain the norm. However, the issue is that we are summing over $n$
elements where $n$ is going to infinity, and extra care is required for
the remainder terms; in particular, we will bound them uniformly over $n$.

It is easy to check that the vector $ \langle{\sqrt{d_i/E}}
 \rangle$ is an
eigenvector of $\tilde{A}$ with eigenvalue one. By virtue of
Assumption~\ref
{assumptionrho} we know that $A$ is connected with high probability,
and so the principal eigenvalue has multiplicity one.
Thus $\tilde{\mathbf{u}}(i)=\sqrt{d_i/E}$. Now termwise Taylor
approximation gives
\begin{equation}
\label{equemp1taylor} \tilde{u}_i = \frac{1}{\sqrt{n}}+
\frac{\bar{d}_i}{2\sqrt
{n(n-1)^2p^2}}+R, 
\end{equation}
where $R$ is a length $n$ vector of remainder terms.
We will now invoke Lemma~\ref{lemnorm-centered-taylor}. Let $c_n$ be
the vector of constants $1/\sqrt{n}$, and $x_n^{1}:=\frac{\bar
{d}_i}{2\sqrt
{d_0E_0}}$, where $d_0$ and $E_0$ are respectively the expectation of
$d_i$ and $E$. In particular $d_0=(n-1)p$ and $E_0=n(n-1)p$.
Hence $\|x_n^{1}\|\sim C_0/\sqrt{n\rho}$, and the mean of $x_n^{1}$
is 
$O_P(1/\sqrt{n^3\rho})=o_P(\|x_n^{1}\|/\sqrt{n})$. Using standard
probabilistic arguments and the form of $R$, we show that $\|R\|=o_P(\|
1/\sqrt{n\rho}\|)$; for details, see Section~D of the Supplement \cite{suppnormalization}.
Hence we have
\[
\|\tilde{\mathbf{r}}\|^2=\sum_i
\biggl(\tilde{u}_i - \sum_i
\tilde{u}_i/n\biggr)^2\sim\frac
{1}{4n(n-1)p^2}\sum
_i \frac{\bar{d}_i^2}{n}.
\]
\upqed\end{pf*}
%
%
%
%
\begin{pf*}{Proof of Corollary~\protect\ref{corerdos-norm-4}}
In order to compute $\tilde{d}_{11}^2$ and $\hat{d}_{11}^2$, we
need to
compute the norms and averages of $\hat{\mathbf{r}}_k$ and
$\tilde{\mathbf{r}}_k,k\in\{1,2\}$
restricted to class $C_1$.
First note that $\hat{\mathbf{r}}_1(C_1)=\hat{\mathbf{r}}_1^T
\mathbf{v}_1/\sqrt{n\pi}=\|\hat
{\mathbf{r}}_1\|^2/\sqrt{n\pi}$ by construction, and $\hat
{r}_2(i)=0$,  for $i\in C_1$. Hence $\sum_{i\in C_1}\hat
{r}_1(i)^2=\|\hat{\mathbf{r}}_1\|^2$.

Using equation~(\ref{eqd11}),
$\hat{d}_{11}^2=(\|\hat{\mathbf{r}}_1\|^2/n\pi-\|\hat{\mathbf
{r}}_1\|^4/n\pi)/\hat
{c}_{11}^2=\|\hat{\mathbf{r}}_1\|^2/n\pi$. But $\|\hat{\mathbf
{r}}_1\|^2$ is the
norm-square of the residual of the principal eigenvector from $A_1$
which is an Erd\H{o}s--R\'{e}nyi $(n\pi,\alpha)$ graph; see
Lemma~\ref{lemr-unnorm}.

Now we consider the corresponding quantities from $\tilde{A}$. The only
issue is that $A$ has two disconnected components (each of which is
connected w.h.p., via Assumption~\ref{assumptionrho}), and hence
$\tilde{A}$ will have two eigenvalues equal to one; hence the
first two
eigenvectors can be any two orthogonal vectors spanning this
eigenspace. Since Euclidean distances (e.g., $\tilde
{d}_{11}^2,\tilde{d}_{12}^2$ etc.) are preserved under rotation,
in this simple setting, any such pair of vectors can be shown to yield the same answer.

We will construct $\mathbf{u}_1$ and $\tilde{\mathbf{u}}_1$ as follows.
$\mathbf{u}_2$ and $\tilde{\mathbf{u}}_2$ are constructed analogously.
\[
\mathbf{u}_1(i)= %
\cases{\displaystyle 1/\sqrt{n\pi}, &
\quad\mbox{for $i\in C_1$},\vspace*{3pt}
\cr
0, &\quad$
\mbox{otherwise}$,} \qquad\tilde{\mathbf{u}}_1(i)= \cases{
\sqrt{d_i/E}, &\quad$\mbox{for $i\in C_1$}$,\vspace
*{3pt}
\cr
0, & $\quad\mbox{otherwise}$.}
\]

Since $\mathbf{u}$ and $\mathbf{v}$ are identical in the zero
communication case, we
have $\tilde{d}_{11}^2=\|\tilde{\mathbf{r}}_1\|^2/n\pi$.
However, $\tilde{\mathbf{r}}_1$ is
simply the residual of the principal eigenvector from~$\tilde
{A}_1$. 
Now an application\vspace*{1pt} of Lemmas~\ref{lemr-unnorm} and~\ref{lemr-norm}
proves equation~(\ref{eqd11-result-nocomm}).

As for $\hat{d}_{12}^2$, note that $\hat{\mathbf{v}}(C_1)=\mathbf
{v}_1^T\hat{\mathbf{v}
}/\sqrt{n\pi}={c}_{11}/\break \sqrt{n\pi}$. Hence, $K_1=\{
{c}_{11}/ \sqrt{n\pi},0\}$ and $K_2=\{{c}_{22}/\sqrt{n(1-\pi
)},0\}$.
Thus $\|K_1-K_2\|^2\sim1/n\pi(1-\pi)$, since both $
{c}_{11}^2=1-\hat{\mathbf{r}}
_1^T\hat{\mathbf{r}}_1$ and ${c}_{22}^2=1-\hat{\mathbf
{r}}_2^T\hat{\mathbf{r}}_2$ are $1-o_P(1)$
(Lemma~\ref{lemr-unnorm}). Since $\hat{d}_{11}^2=O_P(1/n^2\rho)$ is\vspace*{1pt} of
smaller order than $\|K_1-K_2\|^2$, using equation~(\ref{eqd12}) we see
that $\hat{d}_{12}^2\sim1/n\pi(1-\pi)$. An identical argument shows
that $\tilde{d}_{12}^2\sim1/n\pi(1-\pi)$, yielding
equation~(\ref{eqd12-result-nocomm}).
With or without normalization, we have $d_{11}^2=O_P(1/n^2\rho)$,
whereas $d_{12}^2\sim1/n\pi(1-\pi)$; this yields equations~(\ref
{eqd11-d12-ratio-nocomm-unnorm})
and~(\ref{eqd11-d12-ratio-nocomm-norm}). Finally, an identical
argument proves
the result for the normalized and unnormalized versions of $d_{22}^2$
and $d_{21}^2$.
%
\end{pf*}
\section{Analysis of the unnormalized method}
\label{secmainres}
In this section we obtain expressions for $d_{11}^2$ and $d_{12}^2$
when $\gamma\neq0$ for $A$.
First we give a simple lemma describing the eigen-structure of the
conditional probability matrix $P$. The proof is simple and is deferred
to the Supplement.

\begin{lem}
\label{lempopEigVec}
Define a stochastic blockmodel (see Definition~\ref{defSBM}) with
parameters $(\alpha,\beta,\gamma,Z)$, where $\gamma>0$ and $\alpha
\beta
\neq\gamma^2$. 
The two population eigenvectors of~$P$ are piecewise constant with
first $n\pi$ elements $x_1$ and $x_2$, respectively, and the second
$n(1-\pi)$ elements $y_1$ and $y_2$, respectively. 
These elements are of the form $C_0/\sqrt{n}$, and they satisfy the
following:
%
\begin{equation}
\label{eqsomecomm-eig-2}\hspace*{8pt} x_1^2+x_2^2 = 1/n
\pi; \qquad y_1^2+y_2^2 = 1/n(1-
\pi);\qquad x_1y_1+x_2y_2=0.
\end{equation}
%
The two principal population eigenvalues $\lambda_1$ and $\lambda_2$
are of the form $C' n\rho$ and $C''n\rho$, where $C'$ and $C''$ are
deterministic constants asymptotically independent of $n$; also,
$|\lambda_1-\lambda_2|$ is of the form $C''' n\rho$ for some arbitrary
constant $C'''$, when $\gamma>0$.
All other eigenvalues of $P$ are $O(\rho)$.
\end{lem}

We will now lay the groundwork for our result on $\hat{d}_{11}^2$ and
$\hat{d}_{12}^2$. In order to extend the simple zero-communication case
to the general case, we will need some key results, which are listed
below. Recall the following decomposition of the population eigenvector:
\begin{equation}
\label{eqv-r-decomp} \mathbf{v}_k=c_{kk} \hat{
\mathbf{v}}_k + \hat{\mathbf{r}}_k.
\end{equation}
We will need the following three key components in order to use the
same technique as in Lemma~\ref{lemr-unnorm}:
\begin{longlist}[(3)]
\item[(1)] Sharp deviation of empirical eigenvalues. For $\gamma=0$, we
have $\hat{\lambda}_k = \lambda_k+O_P(1)$.
\item[(2)] Upper bound on $\|A\hat{\mathbf{r}}_k\|$. For $\gamma
=0$, we have $\|A\hat{\mathbf{r}}
_k\|=O_P(1)$.
\item[(3)] Bound on the average of $\hat{\mathbf{r}}_k$ restricted
to $C_1$. For
$\gamma=0$ we have $\hat{\mathbf{r}}_k(C_1)=O_P(1/n^{3/2}\rho)$.
\end{longlist}
In Section~E of the Supplement \cite{suppnormalization} we will provide detailed
proofs of the
following theorems, which show that the above results are also true
when $\gamma\neq0$.

In the following lemma we establish a sharp eigenvalue deviation result
for blockmodels similar to the one for Erd\H{o}s--R\'{e}nyi graphs
presented in~\cite{furedikomlos81}.
F\"{u}redi and Koml\'{o}s \cite{furedikomlos81} use the von Mises
iteration (also popularly known as power iteration), which intuitively
returns a good approximation of the principal eigenvalue in a few
iterations if the second eigenvalue is much smaller than the first.
In \cite{furedikomlos81} the second largest eigenvalue of the
adjacency matrix is shown to be an order smaller than the first; hence
two steps of power iteration can be shown to give a $O_P(1)$ close
approximation of $\hat{\lambda}_1$. On the other hand, this
approximation can also be shown to be $O_P(1)$ close to the population
eigenvalue $\lambda_1$, thus giving the sharp deviation bound.

In a stochastic blockmodel the second largest eigenvalue is of the same
order as the
first, which is problematic. However, the third largest eigenvalue can
be shown to be $O_P(\sqrt{n\rho\log n})$. Therefore we design a
two-dimensional von Mises-style iteration argument, so that at any
step, the residual vector is orthogonal to the first two empirical
eigenvectors, and thus a $O_P(1)$ deviation of the empirical
eigenvalues from their population counterparts can be proved. While we
prove this result only for the two class blockmodels, the proof can be
extended easily to $k$-class blockmodels, as long as $\lambda_i,i\in\{
1,\dots,k\}$ are distinct.

\begin{lem}
\label{lemlowcommeigvalbound}
Consider an $n$ node network generated from the semi-sparse stochastic
blockmodel $(\alpha,\beta,\gamma,Z)$ with $\gamma>0$.
We have
\[
\mbox{For }i\in\{1,2\}, \qquad \hat{\lambda}_i =
\lambda_i + O_P(1).
\]
\end{lem}

Next we need to show that $\|A\hat{\mathbf{r}}_k\|=O_P(1),k\in\{1,2\}
$, even when
$\gamma\neq0$.
For definiteness let $k=1$. We want to emphasize that proving $\|\hat
{\mathbf{r}}_1\|
=O_P(1/\sqrt{n\rho})$ is not enough to get the above. By construction
$\hat{\mathbf{r}}_1$ is orthogonal to $\hat{\mathbf{v}}_1$, and
hence $\|A\hat{\mathbf{r}}_1\|$ can be
upper bounded by $|\hat{\lambda}_2| \|\hat{\mathbf{r}}_1\|$. For an
Erd\H{o}s--R\'{e}nyi graph, $\hat
{\lambda}_2=O_P(\sqrt{n\rho})$ leading to an $O_P(1)$ bound, whereas
for a stochastic blockmodel, $\hat{\lambda}_2=O_P(n\rho)$ leading to
a $O_P(\sqrt{n\rho
})$ bound. We show the required result by proving that $\hat{\mathbf
{v}}_2^T\hat{\mathbf{r}}
_1=O_P(1/n\rho)$. Since $\hat{\mathbf{v}}_1$ is orthogonal to $\hat
{\mathbf{v}}_2$,
$\hat{\mathbf{v}}_2^T\hat{\mathbf{r}}_1=\hat{\mathbf{v}}_2^T
\mathbf{v}_1$, which we prove to be
$O_P(1/n\rho)$ in the following lemma.

\begin{lem}
\label{lemboundc11etc}
For the stochastic blockmodel $(\alpha,\beta,\gamma,Z)$ with $\gamma
>0$, 
define $c_{12}:=\mathbf{v}_1^T\hat{\mathbf{v}}_2$ and
$c_{21}:=\mathbf{v}_2^T\hat{\mathbf{v}}_1$.
We have
\begin{eqnarray*}
\|\hat{\mathbf{r}}_1\|^2&=& 1-c_{11}^2=O_P(1/n
\rho), \qquad \|\hat {\mathbf{r}}_2\|^2 =
1-c_{22}^2=O_P(1/n\rho),
\\
c_{12} &:=& \mathbf{v}_1^T\hat{
\mathbf{v}}_2=O_P(1/n\rho), \qquad c_{21} :=
\mathbf{v}_2^T\hat{\mathbf{v} }_1=O_P(1/n
\rho).
\end{eqnarray*}
\end{lem}

The final task is to show that $\hat{r}_k(C_1)$ and $\hat{r}_k(C_2)$
are small. The Cauchy--Schwarz inequality gives $|\hat{r}_k(C_1)|\leq
\|
\hat{\mathbf{r}}_k\|/\sqrt{n}=O_P(1/n\sqrt{\rho})$. However, for a
stochastic blockmodel, by virtue
of stochastic equivalence, $\mathbf{v}_k$ for $k\in\{1,2\}$ is piecewise
constant; that is, all entries in $C_1$ have value $x_k$, whereas all
in $C_2$ have value $y_k$.
Now entries of $\hat{\mathbf{v}}_k$ in $C_1$ ($C_2$) constitute a noisy
estimate of $x_k$ ($y_k$). However, one should be able to get an even
better estimate by considering $\hat{v}_k(C_1)$ and $\hat{v}_k(C_2)$.
Since $\hat{r}_k(C_1)$ reflects the error of $\hat{v}_k(C_1)$ around
$x_k$, it is plausible that $\hat{r}_k(C_1)$ is an order smaller than
$\|\hat{\mathbf{r}}_k\|$, which is what we prove in the following lemma.

\begin{lem}
\label{lemlowcommrmeanbound}
Write $\mathbf{v}_i:=c_{ii}\hat{\mathbf{v}}_i+\hat{\mathbf{r}}_i$
for $i\in\{1,2\}$. Now we have
\[
\mbox{For $i,j\in\{1,2\}$,}\qquad\hat{r}_i(C_j)=O_p
\bigl(1/n^{3/2}\rho\bigr). 
\]
\end{lem}

Before proceeding to prove our main result, we present the following
simple concentration results, which are derived in the Supplement \cite{suppnormalization}.

\begin{lem}
\label{lemsumdcSQ}
Denote $\bar{d}_i^{(1)}$ and $\bar{d}_i^{(2)}$ as the centered degree
of node $i$ restricted
to blocks $C_1$ and $C_2$, respectively. In particular, for $k\in\{1,2\}
$, $\bar{d}^{(k)}_i=\sum_{j\in C_k}(A_{ij}-E[A_{ij}|Z])$.
We have
\begin{eqnarray}
\sum_{i\in C_1} \bigl(\bar{d}_i^{(1)}
\bigr)^2 &\sim& (n\pi)^2\alpha(1-\alpha ),
\nonumber
\\[-8pt]
\label{eqsumdcSQ}
\\[-8pt]
\nonumber
\sum
_{i\in C_1} \bigl(\bar{d}_i^{(2)}
\bigr)^2 &\sim &  n^2\pi(1-\pi)\gamma(1-\gamma),
\\
\label{eqdcoplusdctSQ} \sum_{i\in C_1}\bigl(x_1
\bar{d}_i^{(1)}+y_1\bar{d}_i^{(2)}
\bigr)^2 &\sim& \biggl(x_1^2\sum
_{i\in C_1}\bigl(\bar{d}_i^{(1)}
\bigr)^2+y_1^2\sum
_{i\in
C_1}\bigl(\bar{d}_i^{(2)}
\bigr)^2 \biggr).
\end{eqnarray}
%
\end{lem}



Now we prove Theorem~\ref{thmlocommdistunnorm}. Surprisingly, $\hat
{d}_{12}^2$ can be shown to be $(1+o_P(1))/n\pi(1-\pi)$, which does not
depend on the parameters $\alpha,\beta$ or $\gamma$.
\subsection{Derivation of the distance formulas for the unnormalized method}
We now prove Theorem~\ref{thmlocommdistunnorm}.
\begin{pf*}{Proof of Theorem~\ref{thmlocommdistunnorm}}
We will first prove equation~(\ref{eqlowcomm-final-d11-bound}) and then
equation~(\ref{eqlowcomm-final-d12-bound}).
\begin{longlist}[()]
\item[\textit{Derivation of $\hat{d}_{11}^2$
\textup{[}equation~\textup{(}\protect\ref{eqlowcomm-final-d11-bound}\textup{)}\textup{]}}.]
Define $\hat{\mathbf{r}}_i$ as in equation~(\ref{eqv-r-decomp}).
First note that
$\|\hat{\mathbf{r}}_i\|^2=O_P(1/n\rho)$ by Lemma~\ref
{lemlowcommeigvalbound}.
An argument similar to Lem\-ma~\ref{lemr-unnorm} gives
%
\begin{equation}
\label{eqtwoD-r-form} 
\mbox{For $i\in\{1,2\}$},\qquad(A-\hat{
\lambda}_iI)\hat{\mathbf {r}}_i =(A-P)\mathbf{v}
_i+(\lambda_i-\hat{\lambda}_i)
\mathbf{v}_i.
\end{equation}

As discussed earlier, we have $\hat{\mathbf{r}}_1^T \hat{\mathbf
{v}}_2=\mathbf{v}_1^T \hat{\mathbf{v}}_2$
since $\hat{\mathbf{v}}_1\perp\hat{\mathbf{v}}_2$. But from
Lem\-ma~\ref
{lemlowcommeigvalbound}, we know that $c_{12}=O_P(1/n\rho)$, and
hence the projection of $\hat{\mathbf{r}}_1$ on the second
eigen-space $\hat{\mathbf{v}
}_2\hat{\mathbf{v}}_2^T$ only contributes $\|\hat{\lambda}_2 c_{12}
\hat{\mathbf{v}
}_2\|=O_P(1)$. As $\hat{\lambda}_3=O_P(\sqrt{n\rho\log n})$, $\|
A\hat{\mathbf{r}}_1\|
=O_P(1)$.

We compute $\hat{d}_{11}^2$ by deriving asymptotic expressions of
$1/n\pi\sum_{i\in C_1} \hat{\mathbf{r}}_k(i)^2 - \hat{\mathbf
{r}}_k(C_1)^2$, $k\in\{1,2\}$.
First we show that the second term is of lower order than the first.
This is because $\sum_{i\in C_1} \hat{r}_1(i)^2/n\pi\leq\|\hat
{\mathbf{r}}_1\|
^2/n\pi=O_P(1/n^2\rho)$, but $\hat{\mathbf
{r}}_1(C_1)^2=O_P(1/n^3\rho^2)$ using
Lemma~\ref{lemlowcommrmeanbound}. We will now focus on the elements
of $\hat{\mathbf{r}}_1$ belonging to $C_1$. 
We also denote by $\hat{\mathbf{r}}_1(1)$ the subset of $\hat
{\mathbf{r}}_1$ indexed by nodes in
$C_1$, and thus by $[A\hat{\mathbf{r}}_1](1)$ the subset of vector
$A\hat{\mathbf{r}}_1$ indexed
by $C_1$. Also note that $\|[A\hat{\mathbf{r}}_1](1)\|^2\leq\|A\hat
{\mathbf{r}}_1\|^2=O_P(1)$
\begin{eqnarray*}
[A\hat{\mathbf{r}}_1](1)-\hat{\lambda}_1 \hat{
\mathbf{r}}_1(1)&=& \bigl[(A-P)v_1\bigr](1)+(
\lambda_1-\hat {\lambda}_1)v_1(1),
\\
\sum_{i\in C_1}
\hat{r}_1(i)^2&\sim& \frac{\sum_{i\in C_1}(x_1\bar
{d}_i^{(1)}
+y_1\bar{d}_i^{(2)})^2}{\lambda_1^2}.
\end{eqnarray*}
The last step is valid because $\|(A-P)v_1\|$ can be shown to be
$O_P(\sqrt{n\rho})$ (see Section~A of the Supplement \cite{suppnormalization})
whereas $\|[A\hat{\mathbf{r}}
_1](1)\|=O_P(1)$ and $\|(\lambda_1-\hat{\lambda}_1)v_1(1)\|=O_P(1)$
using Lemma~\ref{lemlowcommeigvalbound}.
Similarly, $\sum_{i\in C_1} \hat{r}_2(i)^2\sim\sum_{i\in
C_1}(x_2\bar{d}_i^{(1)}
+y_2\bar{d}_i^{(2)})^2/\lambda_2^2$. Hence using Lemma~\ref{lemsumdcSQ},
equation~(\ref{eqdcoplusdctSQ}),\vspace*{-2pt} we have
\[
\hat{d}_{11}^2\sim\frac{1}{n\pi} \biggl[ \biggl(
\frac
{x_1^2}{\lambda _1^2}+\frac{x_2^2}{\lambda_2^2} \biggr)\sum_{i\in
C_1}
\bigl(\bar{d}_i^{(1)}\bigr)^2+ \biggl(
\frac {y_1^2}{\lambda_1^2}+\frac
{y_2^2}{\lambda_2^2} \biggr)\sum_{i\in C_1}
\bigl(\bar{d}_i^{(2)} \bigr)^2 \biggr].
\]
Now Lemma~\ref{lemsumdcSQ}, equation~(\ref{eqsumdcSQ}) yields
equation~(\ref{eqlowcomm-final-d11-bound}).

\item[\textit{Derivation of $\hat{d}_{12}^2$
\textup{[}equation~\textup{(}\protect\ref{eqlowcomm-final-d12-bound}\textup{)]}}.]
We recall that equation~(\ref{eqd12}) gives $\hat{d}_{12}^2=\hat
{d}_{11}^2+\|K_1-K_2\|^2$, where $K_k=\{\hat{v}_1(C_k), \hat
{v}_2(C_k)\}
, k\in\{1,2\}$.
From equation~(\ref{eqv-r-decomp}), we see that $\hat
{v}_i(C_1)=(v_i(C_1)-\hat{r}_i(C_1))/c_{ii}$, and hence we have
\begin{eqnarray*}
\hat{v}_i(C_1)-\hat{v}_i(C_2)
&= & \biggl(\frac
{x_i-y_i}{c_{ii}} \biggr)- \biggl(\frac {\hat{r}_i(C_1)-\hat
{r}_i(C_2)}{c_{ii}} \biggr),\qquad
\mbox{$i\in\{1,2\}$}. 
\end{eqnarray*}

We will now show that $\|K_1-K_2\|^2=\sum_{i=1}^2(\hat{v}_i(C_1)-\hat
{v}_i(C_2))^2\sim ((x_1-y_1)^2+(x_2-y_2)^2 )$, which is
$\sim1/n\pi
(1-\pi)$, using Lemma~\ref{lempopEigVec} [equation~(\ref{eqsomecomm-eig-2})].
Since $x_1,y_1,x_2,y_2$ are of the form $C_0/\sqrt{n}$ and
$c_{ii}^2=1-O_P(1/n\rho)$ (Lem\-ma~\ref{lemboundc11etc}), we can show\vspace*{-2pt}
that
\[
\sum_{i=1}^2 \biggl(\frac{x_i-y_i}{c_{ii}}
\biggr)^2=\frac
{1+o_P(1)}{n\pi(1-\pi
)}.
\]
%
Also, for $i\in\{1,2\}$, $\hat{r}_i(C_1)=O_p(1/n^{3/2}\rho)$
(Lemma~\ref{lemlowcommrmeanbound}), and $c_{ii}^2=1-\|\hat{\mathbf
{r}}_i\|
^2=O_P(1/n\rho)$ (Lemma~\ref{lemboundc11etc}), and hence we have
equation~(\ref{eqlowcomm-final-d12-bound}).\quad\qed
\end{longlist}
\noqed\end{pf*}

\section{Analysis of the normalized method}
\label{secsomecomm-norm}
As discussed in Section~\ref{secnocomm}, both $\nu_1$ and $\tilde
{\nu}_1$
(see Table~\ref{tablenotation}) equal one, and $\tilde{\mathbf
{u}}_1(i)=\sqrt
{d_i/E}$. In our analysis what naturally appears is the following
notion of density, defined by the expected degree over $n$. All
expectations are conditioned on $Z$.
Let $\mu_1$ and $\mu_2$ the $E[d_i|Z]/n$ for $i$ in $C_1$ and $C_2$,
respectively. Also let $\mu=\sum_{ij} P_{ij}/n^2$. Hence
$\mu_1:=\pi\alpha+(1-\pi)\gamma-\alpha/n$, and $\mu_2=(1-\pi
)\beta+\pi
\gamma-\beta/n$, and $\mu=\pi\mu_1+(1-\pi)\mu_2$. Also, we
recall that
$\bar{d}_i^{(1)}$ is the centered $d_i^{(1)}$, that is,
$d_i^{(1)}-(n\pi
-1)\alpha$ when $i\in C_1$ and $d_i^{(1)}-n\pi\gamma$ when $i\in C_2$.
The properties of the eigen-spectrum of $\tilde{P}$ are stated in the
following lemma. Its proof is deferred to Section~F of the
Supplement~\cite{suppnormalization}.

\begin{lem}
\label{lempopEigVec-norm}
Define a semi-sparse stochastic blockmodel (see Definition~\ref
{defSBM}) with
parameters $(\alpha,\beta,\gamma,Z)$, where $\alpha\beta\neq
\gamma^2$.
The principal eigenvalues $\nu_1$ and $\nu_2$, and the blockwise
entries $\tilde{x}_1$, $\tilde{y}_1$, $\tilde{x}_2$ and
$\tilde{y}_2$ of the principal
eigenvectors of $\tilde{P}$ are given\vspace*{-3pt} by
%
\begin{eqnarray*}
\nu_1&=&1, \qquad \tilde{x}_1 = \sqrt{
\frac{\mu_1}{n\mu}}, \qquad \tilde{y}_1 =\sqrt{
\frac
{\mu_2}{n\mu}},
\\[-2pt]
\nu_2 &=& 1-\frac{\gamma\mu}{\mu_1\mu_2},\qquad\tilde{x}_2=
\sqrt{\frac{(1-\pi)\mu
_2}{n\pi\mu}},\qquad\tilde{y}_2=-\sqrt{
\frac{\pi\mu
_1}{n(1-\pi)\mu}}.
\end{eqnarray*}
All other eigenvalues of $\tilde{P}$ are $O(1/n)$.
\end{lem}

In order to obtain $\tilde{d}_{11}^2$ [equation~(\ref{eqd11})],
we need
$\sum_{i\in C_1} (\tilde{u}_1(i)-\tilde{u}_1(C_1))^2$. Using
$\tilde{u}(C_1)=\sum_{i\in C_1}\tilde{u}(i)/n\pi$ and arguing as in Lemma~\ref
{lemr-norm}, we
see that
%
\begin{equation}
\label{eqd11-u1} \sum_{i\in C_1} \bigl(
\tilde{u}_1(i)-\tilde{u}_1(C_1)
\bigr)^2\sim \frac{1}{4n^3\mu\mu
_1}\sum_{i\in C_1}
\bar{d}_i^2.
\end{equation}
%
%
Computing $\sum_{i\in C_1} (\tilde{u}_2(i)-\tilde
{u}_2(C_1))^2$ requires more
in-depth analysis, since $\tilde{\mathbf{u}}_2$ cannot be
expressed in closed form
as $\tilde{\mathbf{u}}_1$. Instead we look at a ``good''
approximation of $\tilde{\mathbf{u} }_2$, such that the
approximation error cannot mask its $O_P(1/\sqrt
{n\rho})$ deviation from the population counterpart $\mathbf{u}_2$.
The very
first guess is to construct a vector orthogonal to $\tilde{\mathbf
{u}}_1$. In this
case, we present $\mathbf{u}_g^{0}$ as in equation~(\ref{eqvg}).
Define $E_1:=\sum_{i\in C_1}d_i$ and $E_2:=\sum_{i\in C_2}d_i$ 
\begin{equation}
\label{eqvg} u_g^{0}(i)= %
\cases{
\displaystyle \dfrac{\sqrt{d_i}}{E_1}, & \quad\mbox{for $i\in C_1$},
\vspace*{3pt}
\cr
\displaystyle -\dfrac{\sqrt{d_i}}{E_2},& \quad\mbox{for $i\in
C_2$}.}
\end{equation}
In spite of being a fair guess, $\mathbf{u}_g^0/\|\mathbf{u}_g^0\|$ masks the $O_P(1/\sqrt{n\rho})$
error. So we take a von Mises iteration step starting with $\mathbf{u}_g^0$,
and get a finer approximation, namely $\mathbf{u}_g$. We now present
element-wise Taylor expansions of $\mathbf{u}_g$.

\begin{lem}
\label{lemeig2emptaylor}
Define $\mathbf{u}_g^{0}$ as in equation~(\ref{eqvg}).
We have
\[
\bigl[\tilde{A}\mathbf{u}_g^0\bigr]_i=
\cases{ \displaystyle \dfrac{\nu_2}{n\pi\sqrt{n\mu_1}} \biggl(1-\dfrac
{\bar{d}_i}{2n\mu_1}+
\dfrac{\bar{d}_i^{(1)} }{n\mu_1\nu_2}-\dfrac
{\bar{d}_i^{(2)}}{n\mu_2\nu_2}\dfrac{\pi}{1-\pi}+M_i \biggr),\vspace*{3pt}\cr
\qquad\mbox{$i\in C_1$},\vspace*{3pt}
\cr
\displaystyle -
\dfrac{\nu_2}{n(1-\pi)\sqrt{n\mu_2}} \biggl(1-\dfrac
{\bar{d}_i}{2n\mu_2}-\dfrac {\bar{d}_i^{(1)}}{n\mu_1\nu_2}\dfrac
{1-\pi}{\pi}+
\dfrac{\bar{d}_i^{(2)}}{n\mu_2\nu _2}+M'_i \biggr),\vspace*{3pt}\cr
\qquad\mbox{$i\in
C_2$}.}
\]
The remainder vectors $M$ and $M'$ are of norm $o_P(C_0/\sqrt{\rho})$
\[
\bigl\|\tilde{A}\mathbf{u}_g^0\bigr\|\sim\nu_2
\sqrt{\frac{\mu}{n^2\pi
(1-\pi)\mu_1\mu_2}}.
\]
%
%
\end{lem}

The next lemma shows that $\mathbf{u}_g$ has an approximation error of\break 
$O_P(\sqrt{\log n/n^2\rho^2})$. The proof again is deferred to
Section~F of the Supplement~\cite{suppnormalization}.

\begin{lem}
\label{lemAn-second-eigvec-dev}
Define $\mathbf{u}_g:=\tilde{A}\mathbf{u}_g^{0}/\|\tilde
{A}\mathbf{u}_g^{0}\|$. Let $c_g:=(\tilde{\mathbf{u}
}_2)^T\mathbf{u}_g$, that is, the projection of $\mathbf{u}_g$ on
$\tilde{\mathbf{u}}_2$ and $\mathbf{r}
_g:=\mathbf{u}_g-c_g\tilde{\mathbf{u}}_2$. We have
\[
\|\mathbf{r}_g\|=O_P \biggl(\sqrt{\frac{\log n}{n^2\rho^2}}
\biggr); \qquad c_g=1-o_P(1).
\]
\end{lem}

Now we are ready to derive the expressions of $\tilde{d}_{11}^2$
and $\tilde{d}_{12}^2$ (Theorem~\ref{thmd11-sq-norm}).
\subsection{Derivation of the distance formulas for the normalized method}
We now prove Theorem~\ref{thmd11-sq-norm}.
\begin{pf}
We will first prove equation~(\ref{eqsomecomm-norm-d11SQ}) and then
equation~(\ref{eqsomecomm-norm-d12SQ}).
\begin{longlist}[{}]
\item[\textit{Derivation of $\tilde{d}_{11}^2$ \textup{[}equation~\textup{(}\protect
\ref{eqsomecomm-norm-d11SQ}\textup{)]}}.]
Computing $\tilde{d}_{11}^2$ only involves the entries of~$\tilde{\mathbf{u}}_2$
indexed by nodes in $C_1$; hence we will apply Lemma~\ref
{lemnorm-centered-taylor} on $\tilde{u}_2(i),i\in C_1$.
Using our construction,
%
\begin{equation}
\label{equ-decomp} \tilde{\mathbf{u}}_2=(\mathbf{u}_g-
\mathbf{r}_g)/c_g \qquad \mbox{where
$c_g=1-o_P(1)$.}
\end{equation}
Using Lemma~\ref{lemeig2emptaylor}, for $i\in C_1$, we can write
each term of $\mathbf{u}_g$ as
\[
u_g(i)=\chi_n \bigl(1+ x_n^1(i)+M_i
\bigr),
\]
where $\mathbf{x}^1_n$ and $M$ are the first and remainder terms in
the Taylor
expansion of $u_g(i)/\chi_n$. We have
\begin{eqnarray*}
\chi_n &:=& \dfrac{\nu_2}{n\pi\sqrt{n\mu_1}\|\tilde
{A}\mathbf{u}_g^0\|}\sim\sqrt {
\frac{(1-\pi)\mu_2}{n\pi\mu}},
\nonumber
\\
x_n^1(i)&:=& -\dfrac{\bar{d}_i}{2n\mu_1}+\dfrac{\bar
{d}_i^{(1)}}{n\mu_1\nu_2}-
\dfrac{\bar{d}_i^{(2)}
}{n\mu_2\nu_2}\dfrac{\pi}{1-\pi},\qquad\mbox{$i\in C_1$}.
\end{eqnarray*}
%
We have
\[
\sum_{i\in C_1} \biggl(\frac{u_g(i)-u_g(C_1)}{\chi_n}
\biggr)^2=\sum_{i\in
C_1} \bigl(
\bigl(x^1_n(i)-x^1_n(C_1)
\bigr)+\bigl(M_i-M(C_1)\bigr) \bigr)^2.
\]
While $\|\mathbf{x}_n^1\|=C_0/\sqrt{\rho}$ [Lemma~\ref{lemsumdcSQ},
equation~(\ref{eqdcoplusdctSQ})], $x_n^1(C_1)=O_P(1/\sqrt{n^2\rho})$,
since it involves averages of $O(n^2)$ independent Bernoulli random variables.
Also $\|M\|=o_P(1/\sqrt{\rho})$, and hence using a simple application
of the Cauchy--Schwarz inequality, one has
%
\begin{equation}
\label{equg-leading} \sum_{i\in C_1} \bigl(u_g(i)-u_g(C_1)
\bigr)^2\sim\chi_n^2\sum
_{i\in C_1}x_n^1(i)^2.
\end{equation}
Finally, since $\|\mathbf{r}_g\|^2=O_P(\log n/(n\rho)^2)$ and $\sum_{i\in
C_1} (u_g(i)-u_g(C_1) )^2=C_0/n\rho$, from equations~(\ref
{equ-decomp}) and~(\ref{equg-leading}), we have
\begin{equation}
\label{eqd11-u2} \dfrac{1}{n\pi}\sum_{i\in C_1}\bigl(
\tilde{u}_2(i)-\tilde {u}_2(C_1)
\bigr)^2\sim\dfrac
{\chi_n^2}{n\pi}\sum_{i\in C_1}x_n^1(i)^2.
\end{equation}
%
%
%
With a little algebra, equations~(\ref{eqd11-u1}) and~(\ref{eqd11-u2})
give
\begin{eqnarray*}
\tilde{d}_{11}^2 & \sim&
\frac{1}{n\pi}\sum_{i\in C_1} \biggl(\frac{\mu_1}{n\mu }
\frac{\bar{d}_i^2}{4n^2\mu_1^2}+\dfrac
{(1-\pi)\mu_2}{n\pi\mu} \biggl(-\dfrac{\bar{d}_i}{2n\mu_1}+
\dfrac
{\bar{d}_i^{(1)}}{n\mu_1\nu_2}-\dfrac{\bar{d}_i^{(2)}}{n\mu _2\nu
_2}\dfrac{\pi}{1-\pi} \biggr)^2
\biggr)
\\
&\sim& \frac{1}{n\pi}\sum_{i\in C_1} \biggl[
\frac{(\bar
{d}_i^{(1)})^2}{n^3\pi\mu
_1^2} \biggl(\frac{1}{4}+\frac{(1-\pi)\gamma}{\mu_1\nu_2^2} \biggr)+
\frac{(\bar{d}_i^{(2)}
)^2}{n^3\mu_1^2} \biggl(\frac{1}{4\pi}+\frac{\pi\alpha-\alpha
/n}{(1-\pi)\mu _2\nu_2^2} \biggr) \biggr].
\end{eqnarray*}
The last step uses Lemma~\ref{lemsumdcSQ} [equation~(\ref{eqdcoplusdctSQ})].

\item[\textit{Derivation of $\tilde{d}_{12}^2$
\textup{[}equation~\textup{(}\protect\ref{eqsomecomm-norm-d12SQ}\textup{)]}}.]
Equation~(\ref{eqd12}) gives $\tilde{d}_{12}^2=\tilde
{d}_{11}^2+\|K_1-K_2\|$.
$K_i:=\{\tilde{u}_1(C_i),\tilde{u}_2(C_i)\}$ for $i\in\{1,2\}$.
The Taylor expansion used in Lemma~\ref{lemr-norm} shows that the
second-order terms are $o_P(1/n)$ whereas the first is of the form
$C_0/n$. For $\mu_1\neq\mu_2$, neglecting second-order terms gives
\begin{equation}
\label{eqd12norm-part1} \bigl(\tilde{u}_1(C_1)-
\tilde{u}_1(C_2)\bigr)^2\sim
\frac{(\sqrt{\mu
_1}-\sqrt{\mu
_2})^2}{n\mu}.
\end{equation}
For the second part, equation~(\ref{equ-decomp}) and an argument shown
earlier gives
\begin{eqnarray}
\bigl(\tilde{u}_2(C_1)-\tilde{u}_2(C_2)
\bigr)^2 &\sim& \bigl(u_g(C_1)-u_g(C_2)
\bigr)^2
\nonumber
\\[-8pt]
\label{eqd12norm-part2}
\\[-8pt]
\nonumber
&\sim& \biggl(\sqrt{\frac{(1-\pi)\mu_2}{n\pi\mu}}+\sqrt{\frac
{\pi\mu
_1}{n(1-\pi)\mu}}
\biggr)^2. 
\end{eqnarray}
Putting equations~(\ref{eqd12norm-part1}) and~(\ref{eqd12norm-part2})
together yields equation~(\ref{eqsomecomm-norm-d12SQ}). When $\mu
_1=\mu
_2$, the whole contribution comes from the second eigenvector, and
$\tilde{u}_1$ only contributes $o_P(1/n)$ terms.\quad\qed
\end{longlist}
\noqed\end{pf}
\section{Experiments}
\label{secexp}
We demonstrate the benefit of using normalization via classification
tasks for simulated networks and link prediction experiments for real
world co-authorship networks.
For simulations, we investigate the behavior of misclassification error
with: (a) a fixed parameter setting with increasing $n$ and (b) changing
parameter settings for a fixed $n$. For all simulations, a pair of
training and test graphs are generated from a stochastic blockmodel with
a given
parameter setting. The model is fitted using spectral clustering (with
or without
normalization) using the training graph whereas misclassification error
is computed using the test graph.
%
\subsection{Simulated networks}
%
\begin{figure}[b]
\begin{tabular}{@{}cc@{}}

\includegraphics{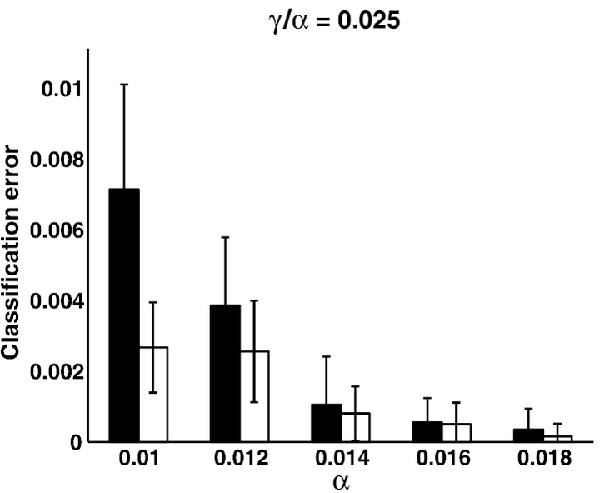}
 & \includegraphics{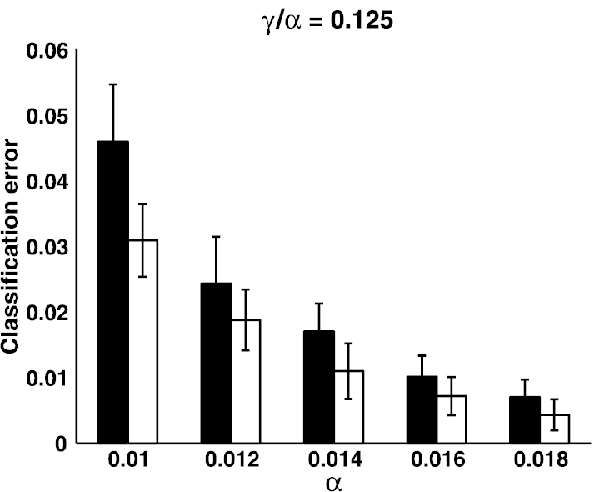}\\
\footnotesize{(A)} & \footnotesize{(B)} \\[3pt]

\includegraphics{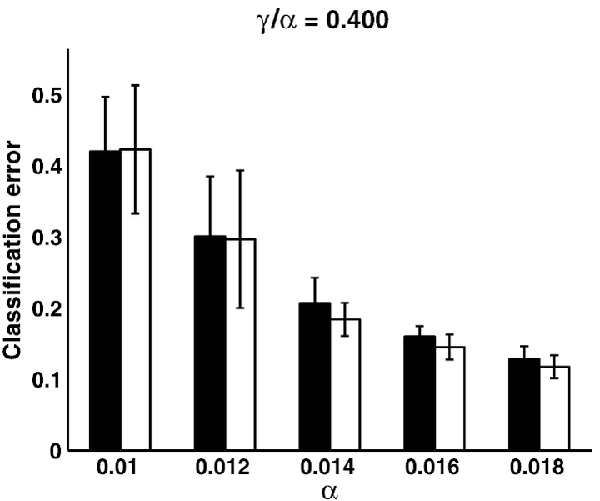}
 & \includegraphics{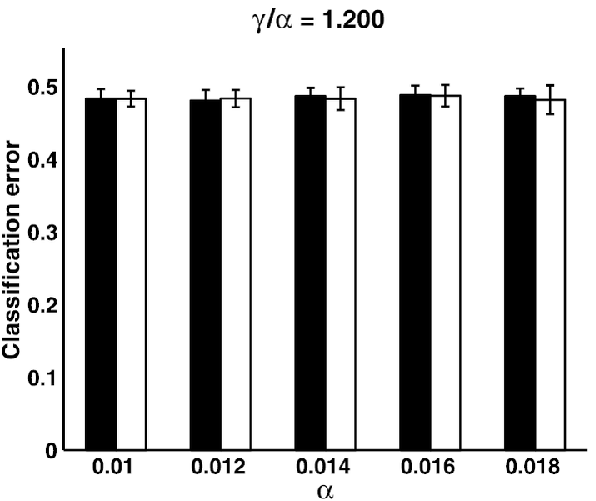}\\
\footnotesize{(C)} & \footnotesize{(D)}
\end{tabular}
\caption{For a fixed $\gamma/\alpha$ ratio
miss-classification error is plotted on the $Y$ axis with increasing
$\alpha$ on the $X$ axis. \textup{(A)} $\gamma/\alpha=0.025$, \textup{(B)} $\gamma
/\alpha
=0.125$, \textup{(C)} $\gamma/\alpha=0.4$ and \textup{(D)} $\gamma/\alpha=1.2$.}
\label{figsimulateVStheory}
\end{figure}

For a stochastic blockmodel with $n=1000$, $\beta=\alpha$ and $\pi
=1/2$, we focus on the
semi-sparse regime, where expected degree is varied from 10--20. We
vary $\alpha\in[0.01,0.018]$ ($y$ axis) and $\gamma/\alpha\in
[0.005,1.2]\setminus\{1\}$ ($x$~axis). The $\gamma/\alpha=1$ case causes
instability because it reduces the stochastic blockmodel to an Erd\H
{o}s--R\'{e}nyi graph and hence is
excluded. Since \texttt{kmeans} can return a local optimum, we run
\texttt{kmeans} five times
and pick the most balanced clustering, in particular the one whose
smallest cluster size is largest among the five runs.

For each of the parameter settings average results from twenty random
runs are reported with error bars.
In order to ensure that our parameter settings reflect the regime of
sparseness required for our theory to hold, we find the connected
components of the graph, and only work with those settings where the
size of the largest connected component is at least 95\% of the size of
the whole graph. All computations are carried out on the largest
connected component. Therefore we never consider the simple case of
disconnected clusters. We also assume that $k=2$ is known.

In each subfigure of Figure~\ref{figsimulateVStheory} we hold $\gamma
/\alpha$ fixed and plot the classification errors of the two algorithms
along the $Y$ axis against increasing $\alpha$ values on the $X$ axis.
Across the subfigures $\gamma/\alpha$ is increased.
Our goal is to turn two knobs to adjust the hardness of the
classification problem. If one increases $\alpha$ for a fixed value of
$\gamma/\alpha$, then the problem becomes easier as the expected degree
increases with increasing~$\alpha$. On the other hand, increasing
$\gamma/\alpha$ makes it hard to distinguish between clusters.

According to our theoretical results, for small $\gamma/\alpha$
ratios, normalization performs better clustering under sparsity. In
Figure~\ref{figsimulateVStheory}(A) and (B), we see that
normalization always has a smaller average error, although the
difference is more striking for small $\alpha$ (average degree about
10). As $\alpha$ is increased, both methods start to perform equally
well. In Figure~\ref{figsimulateVStheory}(C) and (D), $\gamma/\alpha$
is larger, and thus the error rates are also larger. In Figure~\ref{figsimulateVStheory}(C), both methods behave similarly and show
improvement with increasing $\alpha$. Finally in Figure~\ref{figsimulateVStheory}(D), both misclassify about half of the nodes
since the networks become close to Erd\H{o}s--R\'{e}nyi graphs;
possibly with more data
both methods would perform better.
%
\begin{figure}

\includegraphics{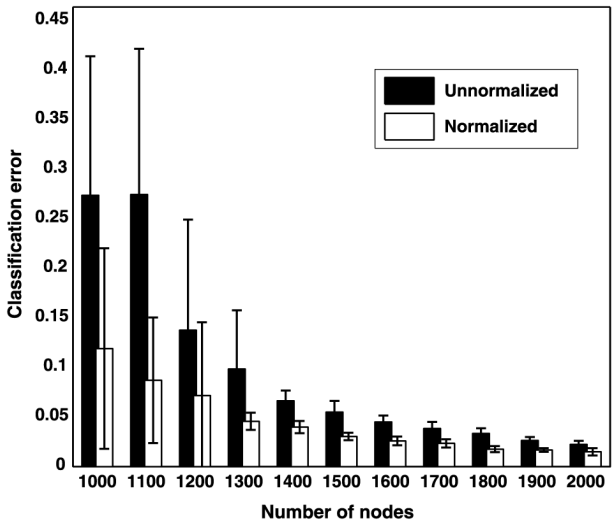}

\caption{Miss-classification error on the $y$ axis and
increasing $n$ on the $x$ axis.}\label{figsim}
\end{figure}
For the second simulation we fix $\alpha=\beta=0.01$, $\gamma=0.002$,
$\pi=0.40$. Now in Figure~\ref{figsim} we plot the error bars on
classification error from twenty random runs along the $Y$ axis, and $n$
is varied from 1000 to 2000 in the $X$ axis. One can see that the
normalized method consistently outperforms the unnormalized method, the
margin of improvement being smaller for $n$ (smaller average degree and
hence sparser graphs).

\subsection{Real world networks}
%
%
For real world datasets we use co-authorship networks over $T$
timesteps. The nodes represent authors, and edges arise if two authors
have co-authored a paper together. Since these networks are unlabeled,
we cannot use classification accuracy to measure the quality of
spectral clustering.
Instead, we choose the task of link prediction to quantitatively assess
the goodness of clustering. Since the number of clusters is unknown, we
learn $k$ via cross validation. We obtain the training graph
($A_1$) by merging the first $T$-2 datasets, use the $T$-1th step ($A_2$)
for cross-validating $k$ and use the last timestep ($A_3$) as the test graph.

We use a subset of the high energy physics (HepTH) co-authorship
dataset ($T=6$), the NIPS data ($T=9$) and the Citeseer data ($T=11$).
Each timestep considers $1$--$2$
years of papers (so that the median degree of the test graph is at
least $1$). 
In order to match the degree regime of our theory, we remove all nodes
with only one neighbor from the training graph, and work with the
largest connected component of the resulting network. Cross validation
and testing are done on the corresponding subgraphs of $T$-1 and $T$th
timesteps, respectively. The number of nodes and average degrees are
reported in Table~\ref{tablereal}.
\begin{table}
\caption{Table of AUC scores for real data}\label{tablereal}
\begin{tabular*}{\tablewidth}{@{\extracolsep{\fill
}}lcccccccc@{}}\hline
&&& \multicolumn{3}{c}{\textbf{AUC scores with
training}}&\multicolumn{3}{c@{}}{\textbf{AUC
scores with training}}\\
&& & \multicolumn{3}{c}{\textbf{links included}}&\multicolumn
{3}{c@{}}{\textbf{links excluded}}\\[-4pt]
&&& \multicolumn{3}{c}{\hrulefill} & \multicolumn
{3}{c@{}}{\hrulefill}\\
\textbf{Dataset} &$\bolds{n}$ &\textbf{Avg. degree}&
\textbf{Unnorm.}&\textbf{Norm.}&\textbf{Katz}&\textbf
{Unnorm.}&\textbf{Norm.}&\textbf{Katz}\\
\hline
HepTH&4795&4.6&0.67&0.82&0.87&0.59&0.79&0.79\\
NIPS&\phantom{0}986&4.4&0.75&0.89&0.75&0.71&0.90&0.69\\
Citeseer&3857&5.6&0.79&0.96&0.97&0.65&0.93&0.90\\
\hline
\end{tabular*}
\end{table}

In Section~\ref{secdisc} we present the misclassification error on
the political blogs network. This is possible because the entities are
labeled as democratic and republican. We preprocess the network as
discussed above, and use $k=2$.
\subsubsection{Link prediction task}
First, we learn the $k\times k$ matrix $\hat{P}$ of within and across
class probabilities by counting edges between (or across) two clusters.
For testing we pick a hundred nodes at random from nodes with at least
one neighbor in the test graph. For node $i$ we construct a prediction
vector of length $n$, whose $j$th entry is the linkage probability
$\hat
{P}_{ab}$ learned using spectral clustering; here node $i$ belongs to
the $a$th
cluster, and node $j$ belongs to the $b$th cluster. For ground truth we
compute the zero one vector representing presence or absence of an edge
between nodes $i$ and $j$ from $A_3$. These vectors are concatenated to
give one prediction vector and the corresponding ground truth.

Now the AUC score of the prediction vector is computed using the
ground truth. This is simply the area under the ROC curve obtained by
plotting the false positive rate along the $x$ axis and the true positive
rate along the $y$ axis. In order to learn $k$, we vary $k$ from ten to a
hundred. For each value of $k$ we estimate $\hat{Z}$ using spectral
clustering with
$k$ eigenvectors of $A_1$ (or its normalized counterpart) and then
estimate the $k\times k$ conditional probability matrix; now AUC scores
are computed using these estimated quantities from $A_2$. The $k$ with
the largest AUC score is picked and mean AUC scores of five random runs
on the test graph using this $k$ is reported.

Since in a co-authorship network, the same edges tend to reappear over
time, it is often possible to achieve high scores simply by predicting
the edges which are already present in the training data. This is why
we examine AUC scores from two experiments: 
%
\begin{longlist}[(2)]
\item[(1)] training links included in the test graph and
\item[(2)] training links excluded from the test graph.
\end{longlist}
The second task is harder.
We compare our methods with the Katz similarity measure between pairs
of nodes \cite{Katz1953}. This measure simply computes a weighted sum
of number of paths between two nodes, the weights decreasing
exponentially as the length of the path grows. It has been shown to
give competitive prediction accuracy for link prediction tasks~\cite
{kleinberg04}. In both panels, the normalized score performs close to
or better than the Katz score, and it outperforms the unnormalized
score consistently.

\section{Summary and discussion}
\label{secdisc}
Normalizing data matrices prior to spectral clustering is a common
practice. In this
paper we propose a theoretical framework to justify this seemingly
heuristic choice. With a series of theoretical arguments, we show that
for a large parameter regime, in the context of network blockmodels,
normalization reduces the variance of points in a given class under the
spectral representation. We also present quantifiable classification
tasks on simulated networks and link prediction tasks on real networks
to demonstrate that normalization improves prediction accuracy. 

While we have only considered two class blockmodels, it should be
possible to generalize our proof techniques to a constant number of
classes if the population eigenvalues for the normalized or the
unnormalized setting are distinct. In order to handle identical
eigenvalues, one would need to update the proof techniques so as to
argue with eigenspaces instead of individual eigenvectors. 
However, our simulations (omitted for brevity) show that a result
similar in flavor to Figure~\ref{figanalytic} would hold. However, as
$\gamma/\alpha$ increases, the ratio grows to one faster compared to
that of the two-class blockmodel.
In fact, for the real world graphs we learn $k$ by cross validation,
and it often exceeds two; our results show that normalization improves
link prediction accuracy in these cases as well.

We conclude this paper with a discussion of some practical
disadvantages of normalization. Unlike $A$, all disconnected components
contribute eigenvalue one to the eigen-spectrum of $\tilde{A}$.
Thus some
of the top eigenvectors of $\tilde{A}$ may have support on a small
disconnected component and may be uninteresting. Another problem
appears in the presence of small subgraphs weakly connected to the rest
of the graph. Here the entries of $\tilde{A}$ corresponding to
edges in the
subgraph may end up having relatively larger values than the rest of
the elements. Hence the top eigenvectors may have high values for nodes
in this subgraph leading to poor clustering.

For concreteness let us consider the political blogs network~\cite
{polblogs}, which is a directed network of hyperlinks connecting nodes
representing weblogs about US politics.\vadjust{\goodbreak} The nodes are labeled as
``liberal'' and ``conservative'' blogs. We symmetrize the network,
remove degree one nodes and find the largest connected component of the
remaining network. On this preprocessed network, misclassification
rates using spectral clustering for the political blogs dataset are 4\% for
normalized versus 37\% for unnormalized.

If the degree one nodes are not removed prior to finding the largest
connected component, then the misclassification error rate is 50\% for
normalized and 40\% for unnormalized. On the other hand, removing
degree-one nodes drastically improves the error rate of the normalized
method to 4\%, while not affecting the unnormalized method's
performance significantly. We have also carried out the link prediction
experiments without removing the degree-one nodes; the relative
behavior of the different algorithms remained essentially unchanged.

In order to alleviate this problem, many regularization
approaches~\cite{chenamini2013,chaudhuri-chung-2012} have
been proposed. 
These approaches ensure that, with high probability, the eigenvalues
corresponding to the discriminating eigenvectors are of larger order
than those corresponding to the uninteresting eigenvectors. Further
analysis of regularization can be found in~\cite{josephyu2013}
and~\cite{qin-rohe2013}.

We want to point out that our analysis is not for a regularized variant
of spectral clustering; however, our experiments do have a
preprocessing step of
operating on the largest connected component after removing low-degree
nodes. This can be thought of as a regularizing procedure since this
often removes small and weakly connected components and ranks the
``useful eigenvectors'' higher. In a nutshell, for the normalized
method, sparse data artifacts may rank uninteresting eigenvectors high.
In this paper we provide theoretical justification for the fact that
the discriminating eigenvectors of $\tilde{A}$ are often more
useful than
those of $A$.

\begin{supplement}[id=suppA]
\stitle{Supplement to ``Role of normalization in spectral
clustering for stochastic blockmodels''}
\slink[doi]{10.1214/14-AOS1285SUPP} 
\sdatatype{.pdf}
\sfilename{aos1285\_supp.pdf}
\sdescription{Because of space constraints we have moved some of the
technical details to the supplementary material~\cite{suppnormalization}.}
\end{supplement}



\printaddresses

\begin{thebibliography}{26}


\bibitem{polblogs}
\begin{binproceedings}[author]
\bauthor{\bsnm{Adamic},~\bfnm{Lada~A.}\binits{L.~A.}} \AND
\bauthor{\bsnm{Glance},~\bfnm{Natalie}\binits{N.}}
(\byear{2005}).
\btitle{The political blogosphere and the 2004 U.S. election: Divided they blog}.
In \bbooktitle{Proceedings of the 3rd Intl. Workshop on Link Discovery}.
\bpublisher{ACM},
\blocation{New York}.
\end{binproceedings}
%

\bptok{imsref}%
\endbibitem

\bibitem{chenamini2013}
\begin{barticle}[mr]
\bauthor{\bsnm{Amini},~\bfnm{Arash~A.}\binits{A.~A.}},
\bauthor{\bsnm{Chen},~\bfnm{Aiyou}\binits{A.}},
\bauthor{\bsnm{Bickel},~\bfnm{Peter~J.}\binits{P.~J.}} \AND
\bauthor{\bsnm{Levina},~\bfnm{Elizaveta}\binits{E.}}
(\byear{2013}).
\btitle{Pseudo-likelihood methods for community detection in large sparse networks}.
\bjournal{Ann. Statist.}
\bvolume{41}
\bpages{2097--2122}.
\bid{doi={10.1214/13-AOS1138}, issn={0090-5364}, mr={3127859}}
\end{barticle}
%

\bptok{imsref}%
\endbibitem

\bibitem{bickel2009nonparametric}
\begin{barticle}[author]
\bauthor{\bsnm{Bickel},~\bfnm{Peter~J.}\binits{P.~J.}} \AND
\bauthor{\bsnm{Chen},~\bfnm{Aiyou}\binits{A.}}
(\byear{2009}).
\btitle{A nonparametric view of network models and Newman Girvan and other modularities}.
\bjournal{Proc. Natl. Acad. Sci. USA}
\bvolume{106}
\bpages{21068--21073}.
\end{barticle}
%

\bptok{imsref}%
\endbibitem

\bibitem{Bol98}
\begin{bbook}[mr]
\bauthor{\bsnm{Bollob{\'a}s},~\bfnm{B{\'e}la}\binits{B.}}
(\byear{1998}).
\btitle{Modern Graph Theory}.
\bseries{Graduate Texts in Mathematics}
\bvolume{184}.
\bpublisher{Springer},
\blocation{New York}.
\bid{doi={10.1007/978-1-4612-0619-4}, mr={1633290}}
\end{bbook}
%

\bptok{imsref}%
\endbibitem

\bibitem{chaudhuri-chung-2012}
\begin{barticle}[author]
\bauthor{\bsnm{Chaudhuri},~\bfnm{Kamalika}\binits{K.}},
\bauthor{\bsnm{Graham},~\bfnm{Fan~Chung}\binits{F.~C.}} \AND
\bauthor{\bsnm{Tsiatas},~\bfnm{Alexander}\binits{A.}}
(\byear{2012}).
\btitle{Spectral clustering of graphs with general degrees in the extended planted partition model.}
\bjournal{Journal of Machine Learning Research---Proceedings Track}
\bvolume{23}
\bpages{35.1--35.23}.
\end{barticle}\vadjust{\goodbreak}
%

\bptok{imsref}%
\endbibitem

\bibitem{Chung-radcliffe}
\begin{barticle}[mr]
\bauthor{\bsnm{Chung},~\bfnm{Fan}\binits{F.}} \AND
\bauthor{\bsnm{Radcliffe},~\bfnm{Mary}\binits{M.}}
(\byear{2011}).
\btitle{On the spectra of general random graphs}.
\bjournal{Electron. J. Combin.}
\bvolume{18}
\bpages{Paper 215, 14}.
\bid{issn={1077-8926}, mr={2853072}}
\end{barticle}
%

\bptok{imsref}%
\endbibitem

\bibitem{donath1973}
\begin{barticle}[mr]
\bauthor{\bsnm{Donath},~\bfnm{W.~E.}\binits{W.~E.}} \AND
\bauthor{\bsnm{Hoffman},~\bfnm{A.~J.}\binits{A.~J.}}
(\byear{1973}).
\btitle{Lower bounds for the partitioning of graphs}.
\bjournal{IBM J. Res. Develop.}
\bvolume{17}
\bpages{420--425}.
\bid{issn={0018-8646}, mr={0329965}}
\end{barticle}
%

\bptok{imsref}%
\endbibitem

\bibitem{feigeofek}
\begin{barticle}[mr]
\bauthor{\bsnm{Feige},~\bfnm{Uriel}\binits{U.}} \AND
\bauthor{\bsnm{Ofek},~\bfnm{Eran}\binits{E.}}
(\byear{2005}).
\btitle{Spectral techniques applied to sparse random graphs}.
\bjournal{Random Structures Algorithms}
\bvolume{27}
\bpages{251--275}.
\bid{doi={10.1002/rsa.20089}, issn={1042-9832}, mr={2155709}}
\end{barticle}
%

\bptok{imsref}%
\endbibitem

\bibitem{Fiedler1973}
\begin{barticle}[mr]
\bauthor{\bsnm{Fiedler},~\bfnm{Miroslav}\binits{M.}}
(\byear{1973}).
\btitle{Algebraic connectivity of graphs}.
\bjournal{Czechoslovak Math. J.}
\bvolume{23}
\bpages{298--305}.
\bid{issn={0011-4642}, mr={0318007}}
\end{barticle}
%

\bptok{imsref}%
\endbibitem

\bibitem{furedikomlos81}
\begin{barticle}[mr]
\bauthor{\bsnm{F{\"u}redi},~\bfnm{Z.}\binits{Z.}} \AND
\bauthor{\bsnm{Koml{\'o}s},~\bfnm{J.}\binits{J.}}
(\byear{1981}).
\btitle{The eigenvalues of random symmetric matrices}.
\bjournal{Combinatorica}
\bvolume{1}
\bpages{233--241}.
\bid{doi={10.1007/BF02579329}, issn={0209-9683}, mr={0637828}}
\end{barticle}
%

\bptok{imsref}%
\endbibitem

\bibitem{HagenK92}
\begin{barticle}[author]
\bauthor{\bsnm{Hagen},~\bfnm{Lars~W.}\binits{L.~W.}} \AND
\bauthor{\bsnm{Kahng},~\bfnm{Andrew~B.}\binits{A.~B.}}
(\byear{1992}).
\btitle{New spectral methods for ratio cut partitioning and clustering.}
\bjournal{IEEE Trans. on CAD of Integrated Circuits and Systems}
\bvolume{11}
\bpages{1074--1085}.
\end{barticle}
%

\bptok{imsref}%
\endbibitem

\bibitem{Hendrickson1995}
\begin{barticle}[mr]
\bauthor{\bsnm{Hendrickson},~\bfnm{Bruce}\binits{B.}} \AND
\bauthor{\bsnm{Leland},~\bfnm{Robert}\binits{R.}}
(\byear{1995}).
\btitle{An improved spectral graph partitioning algorithm for mapping parallel computations}.
\bjournal{SIAM J. Sci. Comput.}
\bvolume{16}
\bpages{452--469}.
\bid{doi={10.1137/0916028}, issn={1064-8275}, mr={1317066}}
\end{barticle}
%

\bptok{imsref}%
\endbibitem

\bibitem{holland-leinhardt}
\begin{barticle}[author]
\bauthor{\bsnm{Holland},~\bfnm{P.~W.}\binits{P.~W.}},
\bauthor{\bsnm{Laskey},~\bfnm{K.~B.}\binits{K.~B.}}
 \AND
\bauthor{\bsnm{Leinhardt},~\bfnm{S.}\binits{S.}}
(\byear{1983}).
\btitle{Stochastic blockmodels: First steps}.
\bjournal{Social Networks}
\bvolume{5}
\bpages{109--137}.
\bid{mr={0718088}}
\end{barticle}
%

\bptok{imsref}%
\endbibitem

\bibitem{josephyu2013}
\begin{barticle}[author]
\bauthor{\bsnm{Joseph},~\bfnm{Antony}\binits{A.}} \AND
\bauthor{\bsnm{Yu},~\bfnm{Bin}\binits{B.}}
(\byear{2013}).
\btitle{Impact of regularization on spectral clustering}.
\bjournal{CoRR}.
\end{barticle}
%

\bptok{imsref}%
\endbibitem

\bibitem{Katz1953}
\begin{barticle}[author]
\bauthor{\bsnm{Katz},~\bfnm{L.}\binits{L.}}
(\byear{1953}).
\btitle{A new status index derived from sociometric analysis}.
\bjournal{Psychometrika}
\bvolume{18}
\bpages{39--43}.
\end{barticle}
%

\bptok{imsref}%
\endbibitem

\bibitem{kleinberg04}
\begin{binproceedings}[author]
\bauthor{\bsnm{Liben-Nowell},~\bfnm{D.}\binits{D.}} \AND
\bauthor{\bsnm{Kleinberg},~\bfnm{J.}\binits{J.}}
(\byear{2003}).
\btitle{The link prediction problem for social networks}.
In \bbooktitle{Conference on Information and Knowledge Management}
\bpublisher{ACM},
\blocation{New York}.
\end{binproceedings}
%

\bptok{imsref}%
\endbibitem

\bibitem{ng01onspectral}
\begin{binproceedings}[author]
\bauthor{\bsnm{Ng},~\bfnm{Andrew~Y.}\binits{A.~Y.}},
\bauthor{\bsnm{Jordan},~\bfnm{Michael~I.}\binits{M.~I.}} \AND
\bauthor{\bsnm{Weiss},~\bfnm{Yair}\binits{Y.}}
(\byear{2001}).
\btitle{On spectral clustering: Analysis and an algorithm}.
In \bbooktitle{Advances in Neural Information
Processing Systems, Vancouver, British Columbia, Canada}.
\bpublisher{MIT Press},
\blocation{Cambridge, MA}.
\end{binproceedings}
%

\bptok{imsref}%
\endbibitem

\bibitem{Oliveira2010}
\begin{bmisc}[author]
\bauthor{\bsnm{Oliveira},~\bfnm{Roberto~Imbuzeiro}\binits{R.~I.}}
(\byear{2009}).
\bhowpublished{Concentration of the adjacency matrix and of the Laplacian in random graphs with independent edges.
Preprint.}
\end{bmisc}
%

\bptok{imsref}%
\endbibitem

\bibitem{Pothen1990}
\begin{barticle}[mr]
\bauthor{\bsnm{Pothen},~\bfnm{Alex}\binits{A.}},
\bauthor{\bsnm{Simon},~\bfnm{Horst~D.}\binits{H.~D.}} \AND
\bauthor{\bsnm{Liou},~\bfnm{Kang-Pu}\binits{K.-P.}}
(\byear{1990}).
\btitle{Partitioning sparse matrices with eigenvectors of graphs}.
\bjournal{SIAM J. Matrix Anal. Appl.}
\bvolume{11}
\bpages{430--452}.
\bid{doi={10.1137/0611030}, issn={0895-4798}, mr={1054210}}
\end{barticle}
%

\bptok{imsref}%
\endbibitem

\bibitem{qin-rohe2013}
\begin{binproceedings}[author]
\bauthor{\bsnm{Qin},~\bfnm{Tai}\binits{T.}} \AND
\bauthor{\bsnm{Rohe},~\bfnm{Karl}\binits{K.}}
(\byear{2013}).
\btitle{Regularized spectral clustering under the degree-corrected stochastic blockmodel}.
In \bbooktitle{Advances in Neural Information Processing Systems, Lake Tahoe, Nevada, USA}.
\bpublisher{MIT Press}, \blocation{Cambridge, MA}.
\end{binproceedings}
%

\bptok{imsref}%
\endbibitem

\bibitem{rohechatterjiyu}
\begin{barticle}[mr]
\bauthor{\bsnm{Rohe},~\bfnm{Karl}\binits{K.}},
\bauthor{\bsnm{Chatterjee},~\bfnm{Sourav}\binits{S.}} \AND
\bauthor{\bsnm{Yu},~\bfnm{Bin}\binits{B.}}
(\byear{2011}).
\btitle{Spectral clustering and the high-dimensional stochastic blockmodel}.
\bjournal{Ann. Statist.}
\bvolume{39}
\bpages{1878--1915}.
\bid{doi={10.1214/11-AOS887}, issn={0090-5364}, mr={2893856}}
\end{barticle}
%

\bptok{imsref}%
\endbibitem


\bibitem{suppnormalization}
\begin{bmisc}[author]
\bauthor{\bsnm{Sarkar},~\bfnm{P.}\binits{P.}} \AND
\bauthor{\bsnm{Bickel},~\bfnm{P.~J.}\binits{P.~J.}}
(\byear{2015}).
\bhowpublished{Supplement to ``Role of normalization in spectral clustering for stochastic blockmodels.''
DOI:\doiurl{10.1214/14-AOS1285SUPP}}.
\bptok{imsref}%
\end{bmisc}

\bptok{imsref}%
\endbibitem

\bibitem{Shimalik}
\begin{barticle}[author]
\bauthor{\bsnm{Shi},~\bfnm{Jianbo}\binits{J.}} \AND
\bauthor{\bsnm{Malik},~\bfnm{Jitendra}\binits{J.}}
(\byear{2000}).
\btitle{Normalized cuts and image segmentation}.
\bjournal{IEEE Trans. Pattern Anal. Mach. Intell.}
\bvolume{22}
\bpages{888--905}.
\end{barticle}
%

\bptok{imsref}%
\endbibitem

\bibitem{priebejasa2012}
\begin{barticle}[mr]
\bauthor{\bsnm{Sussman},~\bfnm{Daniel~L.}\binits{D.~L.}},
\bauthor{\bsnm{Tang},~\bfnm{Minh}\binits{M.}},
\bauthor{\bsnm{Fishkind},~\bfnm{Donniell~E.}\binits{D.~E.}} \AND
\bauthor{\bsnm{Priebe},~\bfnm{Carey~E.}\binits{C.~E.}}
(\byear{2012}).
\btitle{A consistent adjacency spectral embedding for stochastic blockmodel graphs}.
\bjournal{J. Amer. Statist. Assoc.}
\bvolume{107}
\bpages{1119--1128}.
\bid{doi={10.1080/01621459.2012.699795}, issn={0162-1459}, mr={3010899}}
\end{barticle}
%

\bptok{imsref}%
\endbibitem

\bibitem{LuxburgTutorial07}
\begin{barticle}[mr]
\bauthor{\bparticle{von} \bsnm{Luxburg},~\bfnm{Ulrike}\binits{U.}}
(\byear{2007}).
\btitle{A tutorial on spectral clustering}.
\bjournal{Stat. Comput.}
\bvolume{17}
\bpages{395--416}.
\bid{doi={10.1007/s11222-007-9033-z}, issn={0960-3174}, mr={2409803}}
\end{barticle}
%

\bptok{imsref}%
\endbibitem

\bibitem{luxburgConsistency08}
\begin{barticle}[mr]
\bauthor{\bparticle{von} \bsnm{Luxburg},~\bfnm{Ulrike}\binits{U.}},
\bauthor{\bsnm{Belkin},~\bfnm{Mikhail}\binits{M.}} \AND
\bauthor{\bsnm{Bousquet},~\bfnm{Olivier}\binits{O.}}
(\byear{2008}).
\btitle{Consistency of spectral clustering}.
\bjournal{Ann. Statist.}
\bvolume{36}
\bpages{555--586}.
\bid{doi={10.1214/009053607000000640}, issn={0090-5364}, mr={2396807}}
\end{barticle}
%

\bptok{imsref}%
\endbibitem
\end{thebibliography}
\end{document}